\documentclass[10pt,twocolumn,letterpaper]{article}

\usepackage{iccv}
\usepackage{times}
\usepackage{epsfig}
\usepackage{graphicx}
\usepackage{amsmath}
\usepackage{amssymb}
\usepackage{bm}
\usepackage{epstopdf}
\usepackage{booktabs}
\usepackage{multirow}
\usepackage{graphbox}

\usepackage[breaklinks=true,bookmarks=false]{hyperref}

\iccvfinalcopy 

\def\iccvPaperID{} 

\begin{document}

\title{Hierarchical Point-Edge Interaction Network \\ for Point Cloud Semantic Segmentation}

\author{Li Jiang$^{1}$ \quad Hengshuang Zhao$^{1}$ \quad Shu Liu$^{2}$ \quad Xiaoyong Shen$^{2}$ \quad Chi-Wing Fu$^{1}$ \quad Jiaya Jia$^{1,2}$\\
	$^{1}$The Chinese University of Hong Kong \quad $^{2}$Tencent YouTu Lab\\
	{\tt\small \{lijiang, hszhao, cwfu, leojia\}@cse.cuhk.edu.hk} \quad \tt\small \{shawnshuliu, dylanshen\}@tencent.com
}

\maketitle

\begin{abstract}
We achieve 3D semantic scene labeling by exploring semantic relation between each point and its contextual neighbors through edges. Besides an encoder-decoder branch for predicting point labels, we construct an edge branch to hierarchically integrate point features and generate edge features. To incorporate point features in the edge branch, we establish a hierarchical graph framework, where the graph is initialized from a coarse layer and gradually enriched along the point decoding process. For each edge in the final graph, we predict a label to indicate the semantic consistency of the two connected points to enhance point prediction. At different layers, edge features are also fed into the corresponding point module to integrate contextual information for message passing enhancement in local regions. The two branches interact with each other and cooperate in segmentation. Decent experimental results on several 3D semantic labeling datasets demonstrate the effectiveness of our work. 
\end{abstract}
\vspace{-2mm}

\section{Introduction}
With increasing capability of 3D sensing hardware, it is now easy to capture 3D data in many scenarios. Compared with 2D images, 3D data provides richer information about the environment. 3D data is in general view-independent and captures 3D structure, making it possible to incorporate geometry information in scene understanding tasks.

Learning-based approaches~\cite{3dshapenet, 3DSemanticSegmentationWithSubmanifoldSparseConvNet, pointnet, pointnet2, sgpn, voxelnet, pointrcnn} were proposed to solve various 3D vision problems, e.g., shape classification, scene semantic/instance segmentation, and 3D object detection. Unlike 2D images, in which pixel grids are regular with object color information, 3D object data scatters, with most space actually not occupied.
Therefore, directly voxelizing 3D scenes and extending deep neural network operations from 2D to 3D is inefficient. Several voxel-based methods, such as Submanifold Sparse Convolution~\cite{3DSemanticSegmentationWithSubmanifoldSparseConvNet} and O-CNN~\cite{ocnn}, improve the 3D convolution efficiency. However, since voxelization is accompanied by loss of information, high-resolution 3D models are needed to uphold the data precision, even though it unavoidably costs large memory and computation resource.

\begin{figure}
	\begin{center}
		\includegraphics[width=0.99\linewidth]{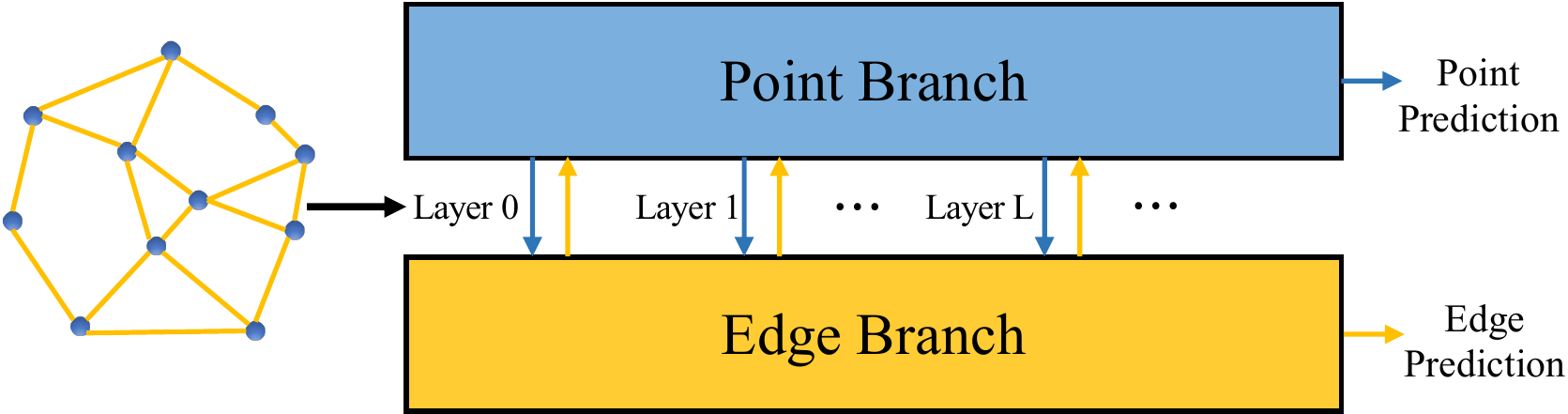}
	\end{center}
	\caption{Simple illustration of our framework. The point and edge branches work together to predict the semantic labels. Self-connected edges and edge directions are omitted.}\vspace{-0.05in}
	\label{fig_simple_framework}
	\vspace{-3mm}
\end{figure}

From another perspective, PointNet~\cite{pointnet} directly processes 3D points in a network, only considering regions covered by the 3D points. PointNet++~\cite{pointnet2} further adopts a hierarchical encoder-decoder structure to consider local regions, which downsamples point clouds in layers first and gradually interpolates them to the original resolution.
This framework just utilizes weak connection between each point and its local context, since point features are extracted independently by the multi-layer perceptrons (MLP).
In segmentation tasks, it is commonly known that local context is crucial for labeling the semantic categories. This motivates us to further explore the semantic relation between points and their local contextual neighbors to extract more discriminative features for 3D semantic scene labeling.

\vspace*{-10pt}
\paragraph{Our Contributions}
To explore the semantic relation between points in a local region and utilize the contextual information, we explicitly build edges between points and their contextual neighbors and establish a hierarchical edge branch with an auxiliary edge loss, as shown in Figure~\ref{fig_simple_framework}.

Specifically, besides the encoder-decoder point branch as in PointNet++, our new edge branch accepts point features from different layers and progressively produces edge features, which are then fed to point branch for fusing information in local graphs. 
For each point, the corresponding edge features provide local intrinsic geometric and regional semantic information to enhance point representation. 

Instead of building isolated graphs for points in each layer, we design a hierarchical graph construction process to gradually take point features at different layers into the edge branch. Edge features of adjacent layers are connected by an operator, named ``edge upsample''.
Consequently, edges on full-resolution point cloud encode multi-layer features, providing comprehensive data for final prediction.

We regularize the final edge features considering semantic consistency of the two connected points, which helps increase the discrimination ability between inter- and intra-category feature pairs, implicitly pulling points with the same semantic label closer in the feature space. 

The decent performance of our method compared with all existing point-based neural networks on the large-scale scene labeling datasets,~\ie, Stanford Large Scale 3D Indoor Space (S3DIS)~\cite{s3dis} and ScanNet~\cite{scannet}, manifests the effectiveness of our framework.

\section{Related Work}

\subsection{3D Representation}
To process 3D data, one typical approach is to store the data in volume grids and adopt 3D convolutions~\cite{3dshapenet, voxnet, volumetric_multiview}.
Since most voxels are unoccupied, Submanifold Sparse Convolution Network~\cite{3DSemanticSegmentationWithSubmanifoldSparseConvNet} defines a sparse convolution operation to process spatially-sparse 3D data.
OctNet~\cite{octnet}, on the other hand, represents the data using unbalanced octrees and defines network operations on these octrees to enable deeper neural networks without sacrificing the precision.
Similarly, O-CNN~\cite{ocnn} uses an octree to enable 3D CNN on high-resolution 3D data.

Another approach is to use multi-view 2D images, to which 2D convolutions~\cite{mvcnn, volumetric_multiview} can be directly applied. However, these approaches overlook the geometric structure in objects and scenes, especially the view-occluded 3D structures. 
Other methods~\cite{tangent_conv, pan2018convolutional} consider 3D object surface and apply convolutions on it for semantic analysis. 

\subsection{Point-based Deep Neural Network}

PointNet~\cite{pointnet} is the first deep neural network to directly process 3D point coordinates, with MLPs and max-pooling for extracting features. Since max-pooling is a global operation on all the points, PointNet lacks local region understanding. 
PointNet++~\cite{pointnet2} further applies a hierarchical structure and uses $k$-NN followed by max-pooling to capture regional information.
Since it aggregates local features simply via a max-pooling, regional information is not yet fully utilized.

Recently, much effort has been made for effective local feature aggregation.
SPLATNet~\cite{splatnet} maps points into a high-dimensional sparse lattice and performs convolution on it. RSNet~\cite{huang2018recurrent} projects features of unordered points into an ordered sequence of feature vectors and applies Recurrent Neural Network
layers to model local dependency. PointCNN~\cite{pointcnn} explores convolution on point clouds and addresses the point ordering issue by permuting and weighting input points and features with the $\mathcal{X}$-Conv operator.
Besides, methods of~\cite{edgefilter, spectralconv, pointconv, dgcnn, continuousconv} explore local context based on graphs.

\begin{figure*}
	\begin{center}
		\includegraphics[width=0.97\linewidth]{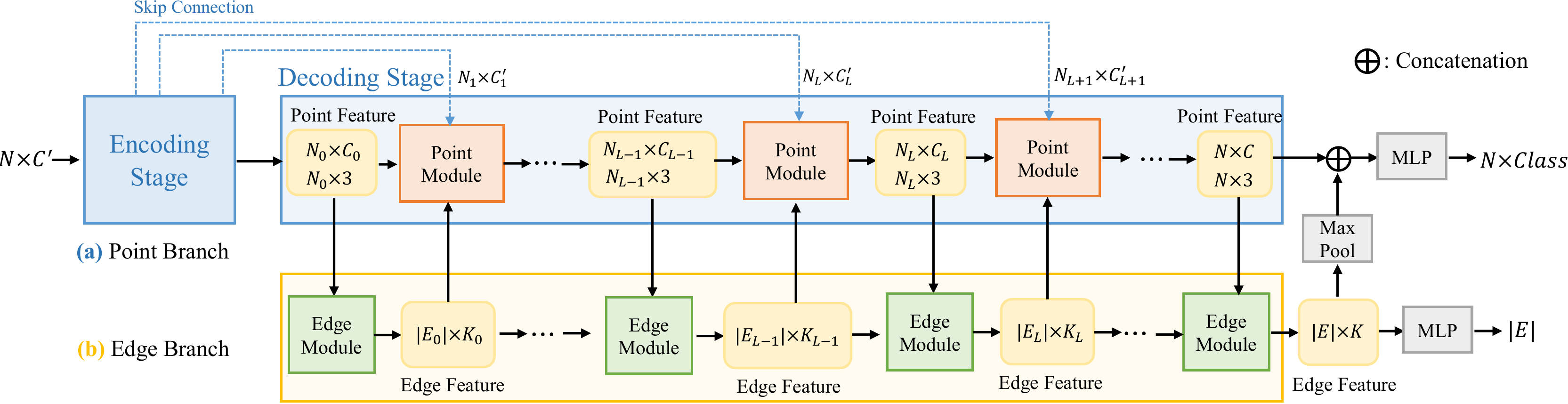}
	\end{center}
	\vspace{-1mm}
	\caption{Overall architecture. $N$ denotes the number of points in the original point cloud. The subscript of $N$ is the layer index. Larger indices indicate layers with more points. $C$ denotes the number of point feature channels. $K$ denotes the number of edge feature channels. $E$ denotes the edge set. The edge feature is encoded from the coarsest layer 0, and is gradually refined with the point features from later layers. Edge features in different layers also participate in the corresponding point modules to provide contextual information. }
	\label{fig_framework}
	\vspace{-1mm}
\end{figure*}

\vspace{-3mm}
\paragraph{Graph-based Methods}
ECC~\cite{edgefilter} organizes point clouds as graphs and uses graph convolutions to dynamically learn weights to combine local features.
DGCNN~\cite{dgcnn} proposes the EdgeConv module to generate edge features that describe the connection between a point and its nearest neighbors.  
PointWeb~\cite{pointweb} further connects every point pairs in a local region to obtain more representative region features. 
KCNet~\cite{kcnet} creates $k$-nearest neighbor graphs and applies kernel correlation to learn local structures over point neighborhood.   
PCCN~\cite{continuousconv} and PointConv~\cite{pointconv} connect each point with its $k$-nearest neighbors and extend the convolution operation from regular grids to irregular point clouds by adaptively projecting the relative position of two points to a convolution weight. Compared to PCCN, PointConv additionally considers point distribution density.
Spectral Graph Convolution~\cite{spectralconv} performs graph convolution after a graph Fourier transform.
Superpoint Graph (SPG)~\cite{spg} splits the point cloud into geometrically-homogeneous partitions and builds a super-point graph, followed by a graph neural network to produce semantic labels.

In our work, we also propose a graph for point cloud processing, and yet focus particularly on exploring the semantic relation between points and their contextual neighbors for semantic segmentation through explicit edges. 
The key distinction of our method from other graph-based frameworks is that instead of fixing the graph and point resolution (e.g., PCCN~\cite{continuousconv} and KCNet~\cite{kcnet}) or building independent graphs at each scale (e.g., PointConv~\cite{pointconv}, PointWeb~\cite{pointweb} and ECC~\cite{edgefilter}), our graph is \emph{hierarchically constructed}. We construct an edge branch, in which we fuse multi-scale point features and propagate edge features over multiple scales to enable longer distances of message passing hierarchically over edges without large memory overhead.
Moreover, we propose edge loss aiming to encode the edges with exact semantic consistency information and increase the discrimination power among point features with different categories. 

With meaningful edge features, we further feed edge features into each scale of the point branch to offer contextual information. To pass messages via edges, PointConv~\cite{pointconv} and PCCN~\cite{continuousconv} adaptively learn weights from edges to fuse point features, while KCNet~\cite{kcnet} defines a point-set kernel and kernel correlation to aggregate local features along edges. Different from these methods, our approach concatenates each point feature with the max-pooled corresponding edge features. Our approach requires less parameters to learn and preserves the distinctiveness of individual point features (Section~\ref{sec_edge_usage} provides more discussions).

\section{Our Approach}

We design a hierarchical edge branch collaborating with the point prediction branch for point cloud semantic segmentation, as shown in Fig.~\ref{fig_framework}. 
We progressively enlarge the graph, upsample edge features, and accept point features in different layers to refine the edge features. Edge features in different layers then provide extra contextual information for point feature learning. The final edge features are regularized with semantic consistency of their two-end points, which serve as auxiliary supervision for point features.  

In this section, we first introduce the new edge branch, covering especially the interaction between point and edge branches, in Section~\ref{sec_edge_branch}.
Then the hierarchical graph construction framework, which enables integration of different-layer information for edge prediction is described in Section~\ref{sec_hierarchical_graph}. Section~\ref{sec_loss} depicts the loss regularizing both category prediction of each point and semantic-consistency prediction of each edge.

\subsection{Edge Branch}
\label{sec_edge_branch}

Given a point cloud with $N$ points $\mathcal{P} = \{p_1, p_2, ..., p_N\}$, we construct a directed graph $G = (V, E)$, where $V = \mathcal{P}$ and $E$ includes the edges that connect each point to its contextual points. Here, $G$ is hierarchically constructed in a coarse to fine manner.
We denote the graph in layer $L$ as $G_L$. A larger $L$ indicates a layer with more points, and layer $0$ is the coarsest layer with the least points. The detailed graph construction process is depicted later in Section~\ref{sec_hierarchical_graph}. Here, we first introduce the constitution of edge branch and how it interacts with the point branch.

As shown in Fig.~\ref{fig_framework}, for the point branch, we follow PointNet++~\cite{pointnet2} to create a hierarchical encoder-decoder structure with previous features in point encoder connected to the corresponding point decoder layers through skip-connection, thus passing detailed low-level information. The point cloud is downsampled and then upsampled in the process. Meanwhile, we construct an edge branch with consecutive edge modules, taking both features from the corresponding point module and the previous edge module. 

\begin{figure}[!t]
	\begin{center}
		\includegraphics[width=0.99\linewidth]{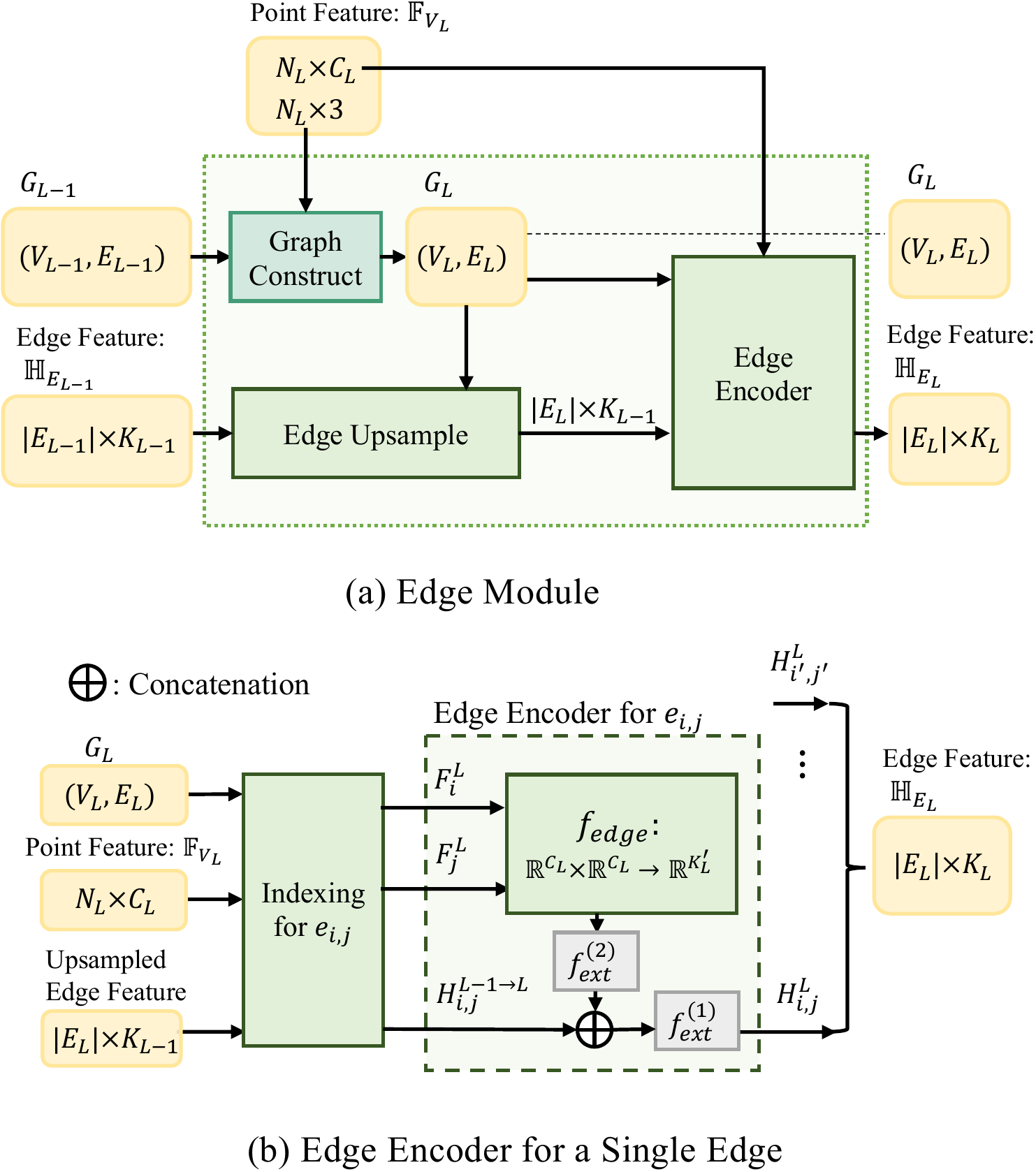}
	\end{center}
	\vspace{-2mm}
	\caption{(a) Architecture of the \textit{Edge Module}. (b) \textit{Edge Encoder} block in (a). $K_L$ and $C_L$ represent the channel numbers in edge and point features in layer $L$, respectively. For simplicity, we only illustrate the edge encoding process for a single edge in (b). Edge features for all the edges in $E_L$ constitute $\mathbb{H}_{E_L}$.}
	\label{fig_edge_module}
	\vspace{-4mm}
\end{figure}

The procedure is to extract edge features from the coarsest layer to grab high-level information with the largest receptive field, and progressively fuse point features from finer layers into edges, in parallel with the point decoding stage. Point features from the encoder layers are also used in the process, along with 
skip-connection to the corresponding decoder layers. 

Although both abstract global features from the coarser layers and detailed information from finer layers are important, the most essential data for edge prediction is from the last layer with the most refined point features. With this consideration, edge features are encoded in a coarse-to-fine manner, making point features in the finest layer fused at last. The hierarchical edge features are also fed to the corresponding point modules to provide additional contextual information.

\subsubsection{Edge Module}
\label{sec_edge_encoder}
At the decoding stage, 
for layer $L$, we denote the graph as $G_L = (V_L, E_L)$ and the number of points as $N_L$.
The edge module accepts the $L$-layer point features $\mathbb{F}_{V_L}$ and $(L-1)$-layer edge features  $\mathbb{H}_{E_{L-1}}$ as arguments and returns the edge features in layer $L$. As shown in Fig.~\ref{fig_edge_module}(a), the edge module is expressed as 
\begin{equation}
\mathbb{H}_{E_L} = M_{encoder}(\mathbb{F}_{V_L}, M_{upsample}(\mathbb{H}_{E_{L-1}})),
\end{equation}
where $M_{encoder}$ denotes the edge encoder and $M_{upsample}$ is the edge upsampling module, which maps edge features in graph $G_{L-1}$ to graph $G_{L}$. The graph construction and edge upsampling process will be described in Section~\ref{sec_hierarchical_graph}.

For each edge $e_{i, j} = (p_i, p_j) \in E_L$, its edge feature at layer $L$ is written as
\begin{equation}
H_{i, j}^{L} = M_{encoder}(F_i^{L}, F_j^{L}, H_{i, j}^{L-1\to L}),
\end{equation}
where $F_i^{L}$ and $F_j^{L}$ are the point features of $p_i$ and $p_j$, respectively.
$H_{i, j}^{L-1\to L}$ is the edge feature upsampled from layer $L-1$ to layer $L$.

As illustrated in Fig.~\ref{fig_edge_module}(b),  $M_{encoder}$ for a single edge can be expanded as
\begin{equation} 
H_{i, j}^{L} = f_{ext}^{(1)}([f_{ext}^{(2)}(f_{edge}(F_i^{L}, F_j^{L})), \text{ }H_{i, j}^{L-1\to L}]),
\end{equation}
where $[\cdot, \cdot]$ represents concatenation.
The feature extractor $f_{ext} : \mathbb{R}^{n} \to \mathbb{R}^{m}$ can be any differentiable function. In our implementation, we apply MLP as $f_{ext}$. The edge function $f_{edge} $ takes the two point features it connects as input and outputs a feature for the edge. We formulate $f_{edge}$ as 
\begin{equation}
f_{edge}(F_i^{L}, F_j^{L}) =   [(p_j - p_i),\text{ } F_j^{L},\text{ }  F_i^{L}] ,
\end{equation}
where $[\cdot, \cdot, \cdot]$ concatenates the three elements, and $p_i, p_j$ here represent 3D point coordinates. The two point features are concatenated for completely preserving information of the two points. Also, we provide $(p_j - p_i)$ to indicate the relative position between the two points. Other implementations of $f_{edge}$ are discussed in the experiment part.

\begin{figure}
	\begin{center}
		\includegraphics[width=0.99\linewidth]{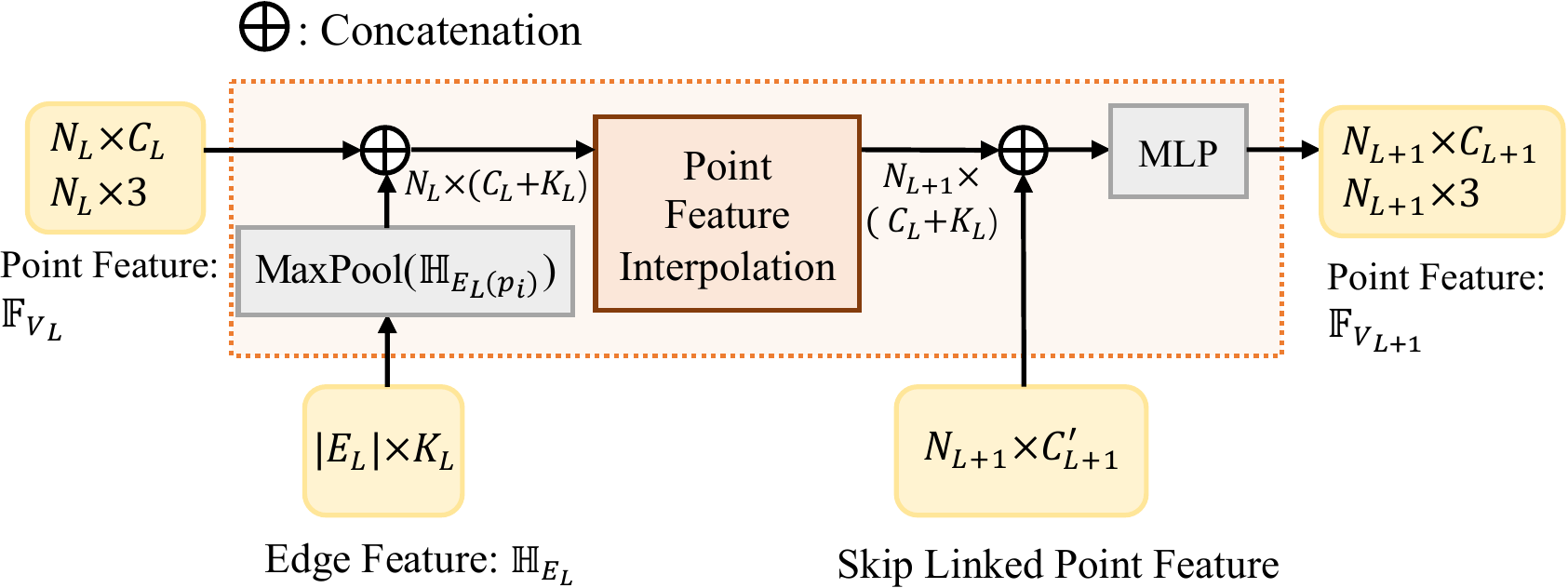}
	\end{center}
	\caption{Architecture of the \textit{Point Module}. $K_L$ denotes the channel number of the $L$-layer edge features, while $C_L$ denotes the channel number of the $L$-layer point feature.}
	\vspace{-2mm}
	\label{fig_point_module}
\end{figure}

\subsubsection{Incorporation of Edges in Point Prediction}
\label{sec_interaction}
For layer $L$, every point in graph $G_L$ links to other contextual points. So corresponding edges are expected to pass the contextual information back to the point. To this end, the edge features with respect to point $p_i$ are operated by max-pooling as a region guidance. Let $E_L(p_i)$ denote the set containing all edges starting from $p_i$, the corresponding set of edge features is
\begin{equation}
\mathbb{H}_{E_L(p_i)} = \{H_{i, j}^{L} | (p_i, p_j) \in E_L(p_i)\}.
\end{equation}
The point feature $F_i^{L}$ is then updated by
\begin{equation}
(F_i^{L})_{new} = [F_i^{L}, \text{ } \mbox{MaxPool}(\mathbb{H}_{E_L(p_i)} )].
\end{equation}
Fig.~\ref{fig_point_module} gives an illustration of the process. 

By incorporating edge information in point features, we enlarge the message passing range. The local region feature provided by the edges allows the point feature extractor to see farther in each layer.
Additional contextual information including intrinsic geometry and semantic relation in the local region is incorporated in the region feature to benefit segmentation.
We experiment with other schemes for message passing. Section~\ref{sec_edge_usage} gives more discussions.

By helping feature extraction in the other branch, point and edge features become more powerful in final prediction.

\subsection{Hierarchical Graph Construction}
\label{sec_hierarchical_graph}
\begin{figure*}
	\begin{center}
		\includegraphics[width=0.85\linewidth]{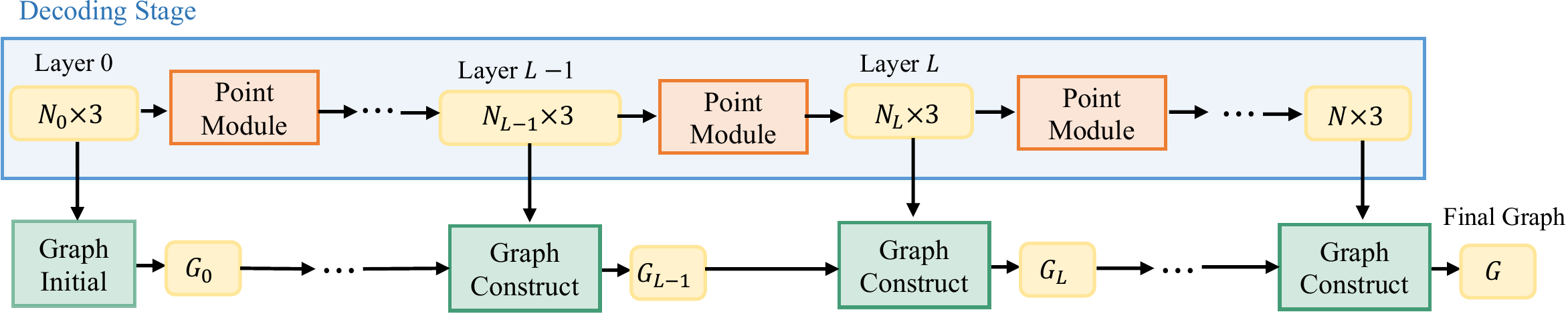}
	\end{center}
	\vspace{-1mm}
	\caption{Hierarchical Graph Construction. The graph is initialized in the coarsest layer and is progressively enlarged by considering both point coordinates in the current layer and the graph in previous layer.}
	\label{fig_graph_construct}
\end{figure*}

Instead of building graphs separately at each layer, we build the graph hierarchically, as shown in Fig.~\ref{fig_graph_construct}. By designing the ``edge upsample'' operation with each edge aware of associated edges in previous layer, we enlarge the receptive field and enable longer-range message passing for edges.

\subsubsection{Graph Initialization}
As shown in Fig.~\ref{fig_graph_construct}, the graph is initialized in the coarsest layer (layer 0). 
The initial graph $G_0$ is constructed by connecting each point with its nearest $k_0$ points. 
Mathematically, $G_0 = (V_0, E_0)$ is formulated as 
\begin{equation} \small
\begin{aligned}
\begin{cases}
V_0 &= \mathcal{P}_0, \\
E_0 &= \{(p_i, p_j) | \text{ } p_i \in \mathcal{P}_0, \text{ } p_j \in N_{k_0}(p_i)\},
\end{cases}
\end{aligned}
\end{equation}
where $\mathcal{P}_0$ is the point set in layer 0, which is downsampled from the original point set with farthest point sampling (FPS) in encoding layers. $N_{k_0}(p_i)$ is the set of the $k_0$-nearest neighbors of point $p_i$, including itself.

\subsubsection{Hierarchical Architecture}
Along with the decoding process of point features, we gradually enlarge the graph and enrich the edge features with more details. The process is illustrated in Fig.~\ref{fig_graph_construct}.
\vspace{-0.12in}
\paragraph{Graph Construction of Layer $L$}
Consider two adjacent layers $L-1$ and $L$ with vertices $V_{L-1}$ and $V_{L}$ as the point set in that layer, respectively.
The graph $G_L$ is constructed by first finding the $k_L$ nearest neighbors for each point in $V_{L}$.
Let $G_{L}^{(0)}=(V_L, E_L^{(0)})$ denote such initial $L$-layer graph. For each edge $e_{i, j} = (p_i, p_j) \in E_L^{(0)}$, we consider the set consisting of possible neighboring edges in layer $L-1$ as
\begin{equation}
E_{ne}^{L-1}(e_{i, j}) = \{(p'_i, p'_j) | \text{ }p'_i \in N_k^{L-1}(p_i),  \text{ }p'_j \in N_k^{L-1}(p_j)\}, \nonumber 
\end{equation}
where $N_k^{L-1}(p_i) \subseteq V_{L-1}$ is the $k$-nearest neighbors of $p_i \in V_L$ in layer $L-1$. $p_i$ is included in $N_k^{L-1}(p_i)$ if $p_i \in V_{L-1}$.

We then check whether edges in $E_{ne}^{L-1}(e_{i, j}) $ exist in $E_{L-1}$ -- the edge set of $G_{L-1}$. If edge $e_{i, j}$ connects two distant points, for which even in the coarser layer $L-1$ there is no connection between the two corresponding regions, we do not take the edge into consideration in layer $L$. Hence, if $E_{ne}^{L-1}(e_{i, j})  \cap E_{L-1} = \O$,  edge $e_{i, j}$ is discarded from $E_L^{(0)}$. Following this principle, the final graph $G_L = (V_L, E_L)$ has an edge set of
\begin{equation} \small
E_L = \bigcup_{p_i \in V_L} E_L(p_i), \nonumber 
\end{equation}
where $E_L(p_i)$ (edges starting from $p_i$) is expressed as
\begin{equation} \small
E_L(p_i) = \{  (p_i, p_j) | p_j \in N_{k_L}(p_i), \text{ }E_{ne}^{L-1}(e_{i, j})  \cap E_{L-1} \neq \O\}.\nonumber 
\end{equation}
Note that at least $e_{i, i}$ is reserved in $E_L(p_i)$ in some extreme cases.

\vspace{-0.12in}
\paragraph{Edge Upsampling}

\begin{figure}
	\begin{center}
		\includegraphics[width=0.99\linewidth]{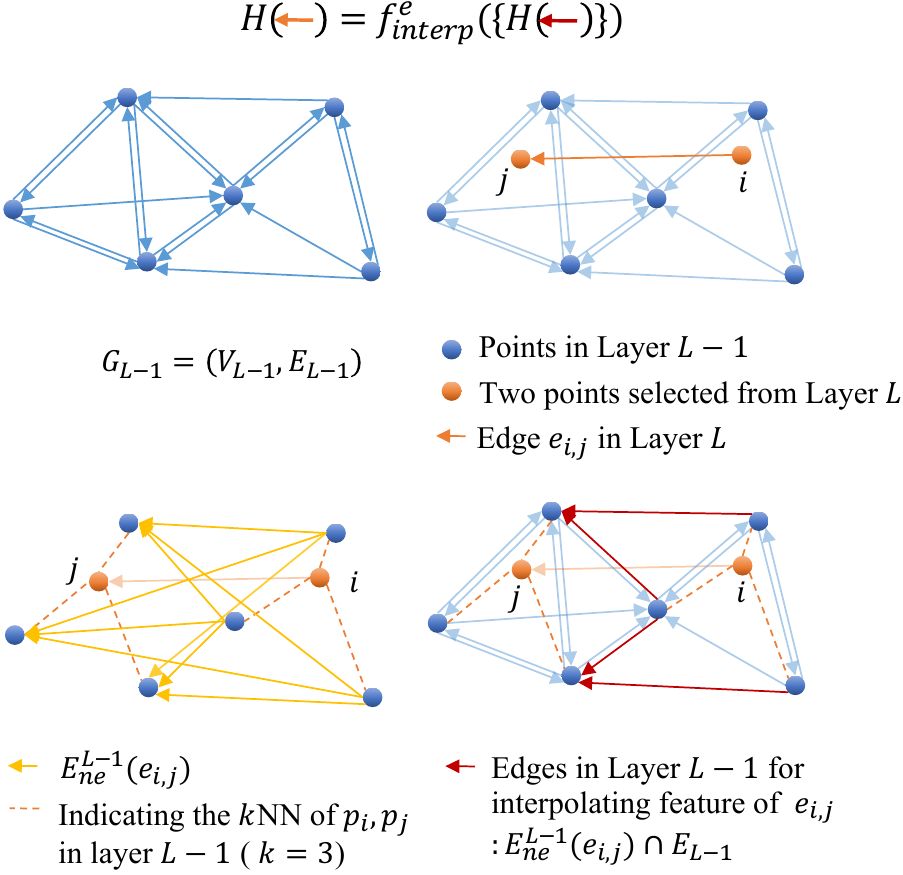}
	\end{center}
	\caption{Demonstration of \textit{edge upsampling}. Points in layer $L-1$ (blue ones) also exist in layer $L$. Self-connected edges are omitted. For edge $e_{i, j}$ in layer $L$, we propagate edge features in layer $L-1$ by finding its neighboring edges in $E_{L-1}$ and interpolating features of these edges. Red arrows represent edges in $G_{L-1}$ for interpolation, which denote intersection of $E_{L-1}$ (blue arrows) and $E_{ne}^{L-1}(e_{i, j})$ (yellow arrows). } 
	\vspace{-3mm}
	\label{fig_edge_upsample}
\end{figure}

In PointNet++~\cite{pointnet2}, point feature of $p_i$ in layer $L$ is propagated from layer $L-1$ by interpolating feature values of its $k$ nearest neighbors in layer $L-1$ as
\begin{equation}
F_i^{L-1 \to L} = f_{interp}^p(\{F_j^{L-1} | \text{ }p_j \in N_k^{L-1}(p_i) \}).
\end{equation}
We similarly propagate edge features in layer $L-1$ to layer $L$ as
\begin{equation}\small
H_{i, j}^{L-1\to L} = f_{interp}^e(\{ H_{i', j'}^{L-1} | \text{ }(p_{i'}, p_{j'}) \in E_{ne}^{L-1}(e_{i, j}) \cap E_{L-1}\}).\nonumber
\end{equation}
A demonstration is given in Fig.~\ref{fig_edge_upsample}.

The interpolation weights are based on the inverse distance of the two pairs of end points. For $H_{i', j'}^{L-1}$, the weight is formulated as
\begin{equation}
w_{i', j'} = \frac{1}{(\| p_i - p_{i'} \|^t + \epsilon)\cdot(\| p_j - p_{j'} \|^t + \epsilon)},
\end{equation}
where $p_{i'}, p_{j'} \in V_{L-1}$, $p_i, p_j \in V_{L}$ represent point coordinates, $\epsilon = 1e-8$ and $t$ is set to 2. The weights are then normalized as
\begin{equation}
\label{eq_normal}
w_{i', j'}^n = \frac{w_{i', j'}}{\sum_{(p_{i''}, p_{j''}) \in E_{ne}^{L-1}(e_{i, j}) \cap E_{L-1}}w_{i'', j''}} .
\end{equation}

\subsection{Loss Function}
\label{sec_loss}
We optimize the point and edge branches jointly with the combined loss on the two branches as
\begin{equation}
L  = \lambda_1 L_{point} + \lambda_2 L_{edge},
\end{equation}
where $\lambda_1$ and $\lambda_2$ adjust the ratio of the two losses.

\begin{table*}
	\begin{center}
	\small
	\resizebox{0.9\linewidth}{!}{%
	\setlength{\tabcolsep}{2.5pt}
		\begin{tabular}{ l | l l l | c c c c c c c c c c c c c}
			\toprule[1.0pt]
			Methods & OA & mAcc & mIoU & ceiling & floor & wall & beam & column & window & door & table & chair & sofa & bookcase & board & clutter \\
			\hline
			PointNet~\cite{pointnet} & - & 48.98 & 41.09 & 88.80 & 97.33 & 69.80 & 0.05 & 3.92 & 46.26 & 10.76 & 58.93 & 52.61 & 5.85 & 40.28 & 26.38 & 33.22\\
			SegCloud~\cite{segcloud} & - & 57.35 & 48.92 & 90.06 & 96.05 & 69.86 & 0.00 & 18.37 & 38.35 & 23.12 & 70.40 & 75.89 & 40.88 & 58.42 & 12.96 & 41.60\\
			PointCNN~\cite{pointcnn} & 85.91 & 63.86 & 57.26 & 92.31 & 98.24 & 79.41 & 0.00 & 17.60 & 22.77 & 62.09 & 74.39 & 80.59 & 31.67 & 66.67 & 62.05 & 56.74\\
			SPGraph~\cite{spg} & 86.38 & 66.50 & 58.04 & 89.35 & 96.87 & 78.12 & 0.00 & 42.81 & 48.93 & 61.58 & 84.66 & 75.41 & 69.84 & 52.60 & 2.10 & 52.22\\
			PCCN~\cite{continuousconv} & - & 67.01 & 58.27 & 92.26 & 96.20 & 75.89 & 0.27 & 5.98 & 69.49 & 63.45 & 66.87 & 65.63 & 47.28 & 68.91 & 59.10 & 46.22\\
			\hline
			Our Method & \textbf{87.18} & \textbf{68.30} & \textbf{61.85} & 91.47 & 98.16 & 81.38 & 0.00 & 23.34 & 65.30 & 40.02 & 75.46 & 87.70 & 58.45 & 67.78 & 65.61 & 49.36 \\
			\bottomrule[1.0pt]
		\end{tabular}
	}
	\end{center}
	\caption{Semantic segmentation results evaluated on S3DIS Area 5. Most methods do not perform well on the ``beam'' category, which has few points (0.029\%).	\label{tab:s3disresult}
	}
\end{table*}

\vspace{-0.12in}
\paragraph{Point Loss}
The final point features are followed by an MLP to produce point-wise semantic predictions. We further use the final edge predictions as weights to aggregate point scores and get refined point predictions. Cross entropy loss is applied to constrain the point predictions.

\vspace{-0.12in}
\paragraph{Edge Loss} 
The edge features in the final graph $G$ are regularized by the edge labels, which represent whether the two-end points of the edge are in the same category or not. The label for edge $e_{i, j} = (p_i, p_j) \in E $ is set as
\begin{equation}
l_{i, j}^{e} = 
\left \{
\begin{array}{lr}
1, &\text{if }l_i^p = l_j^p \\
0, &\text{if } l_i^p \neq l_j^p
\end{array}.
\right.
\end{equation}
where $l_i^p$ and $l_j^p$ are the point semantic labels of $p_i$ and $p_j$.
An MLP is adopted to produce the per-edge prediction. Binary cross entropy loss is chosen for the edge loss as
\begin{equation}\footnotesize
L_{edge} = - \frac{1}{|E|}\sum_{e_{i, j} \in E} ( l_{i, j}^e \log(pred_{i, j}^e) + \alpha (1 - l_{i, j}^e) \log (1 - pred_{i, j}^e)),
\end{equation}
where $pred_{i,j}^e$ is the edge prediction for $e_{i,j}$, and $\alpha$ balances the two kinds of edges, as there are more intra-class edges than inter-class ones considering the local neighborhood. 

The final edge feature for each edge can be deemed as a function on features of the two regions centered at the two-end points. Information from different layers are taken into account.
More details are preserved by encoding at last. Hence, the edge loss guides the edge encoder to seek difference between the intra- and inter-class feature pairs, and implicitly serves as auxiliary supervision for point features. It increases the discrimination power among point features in different categories. Also, with the edge supervision, more exact contextual information is passed to points via edges to enhance point features.

\section{Experiments}
We conducted experiments on two representative and challenging large-scale scene labeling datasets,~\ie, S3DIS~\cite{s3dis} and ScanNet v2~\cite{scannet}, with ablation analysis presented on the ScanNet v2 val set and S3DIS Area 5.

\begin{figure*}[t]
	\centering
	\resizebox{0.99\linewidth}{!}{
	\begin{tabular}{@{\hspace{0.0mm}}c@{\hspace{1.0mm}}c@{\hspace{1.0mm}}c@{\hspace{1.0mm}}c@{\hspace{1.0mm}}c@{\hspace{0.8mm}}c@{\hspace{0.0mm}}}
		\rotatebox[origin=c]{90}{Input} &
		\includegraphics[align=c, height=0.13\linewidth]{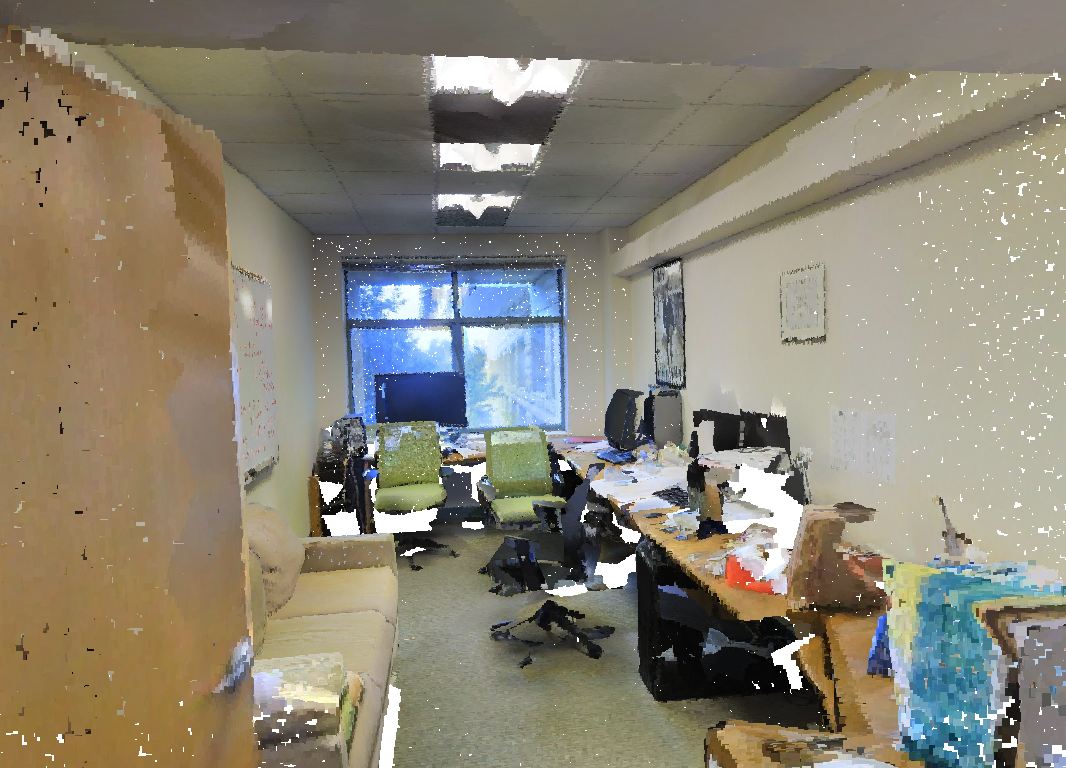}&
		\includegraphics[align=c, height=0.13\linewidth]{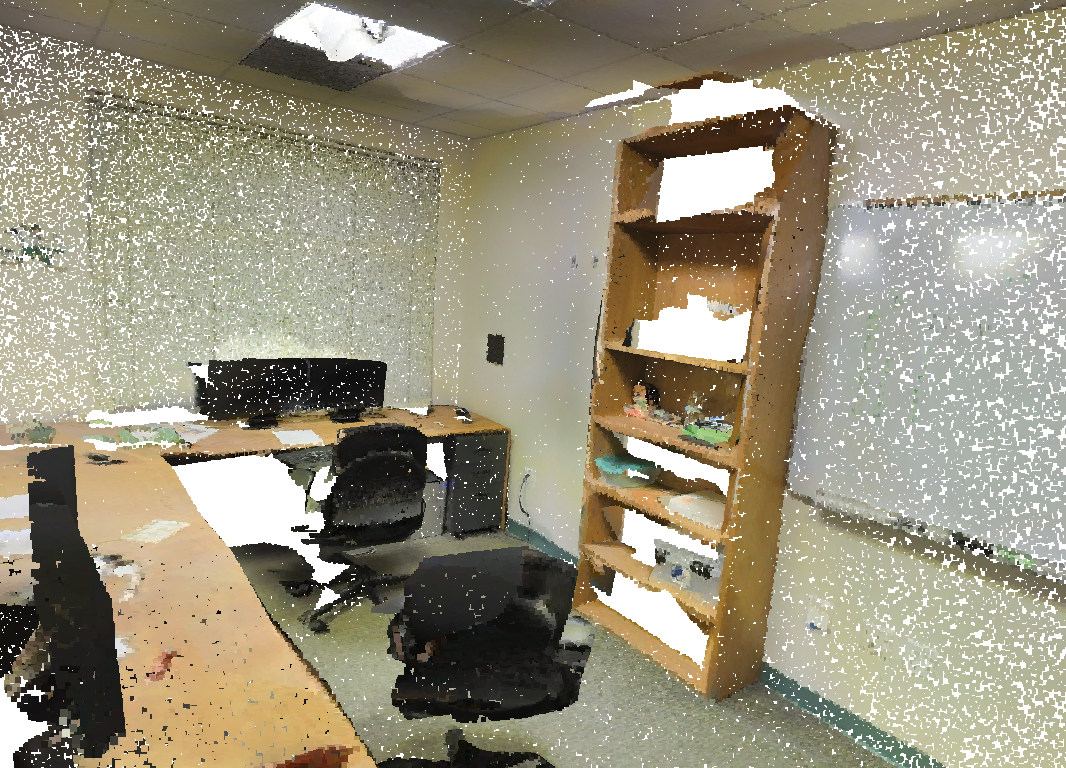}&
		\includegraphics[align=c, width=31mm, height=0.13\linewidth]{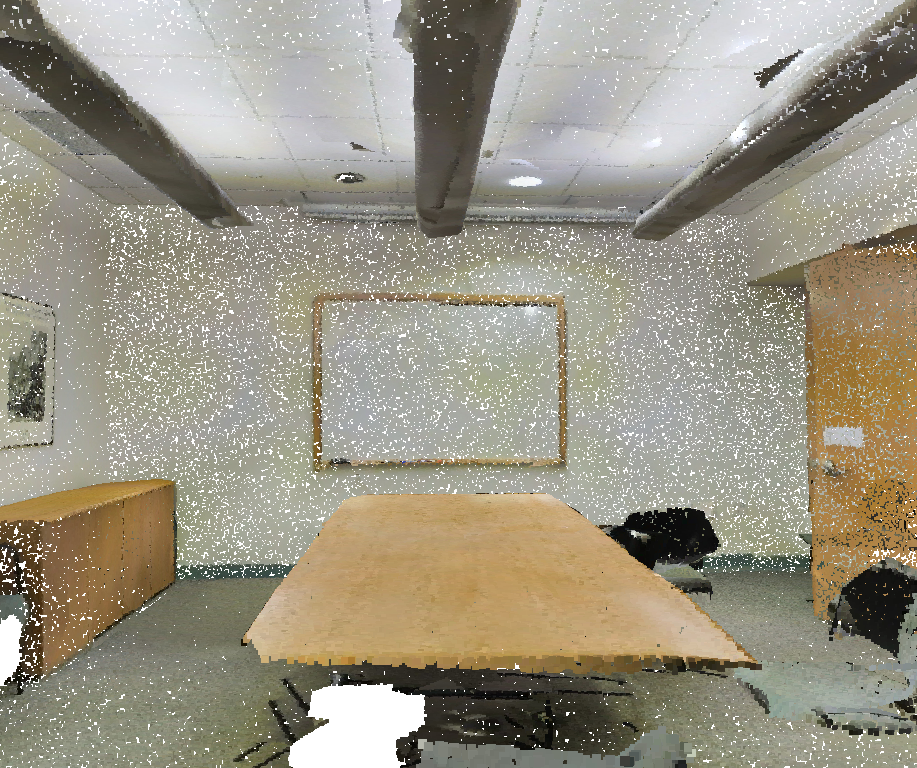}&
		\includegraphics[align=c, height=0.13\linewidth]{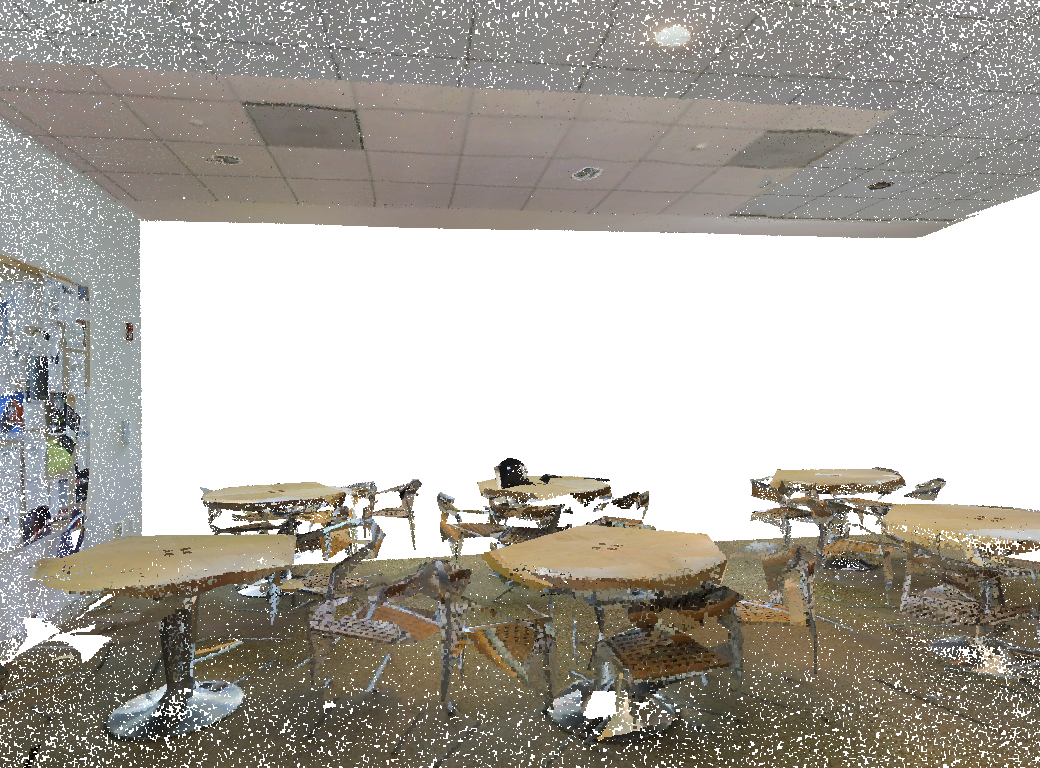}&
		\includegraphics[align=c, height=0.13\linewidth]{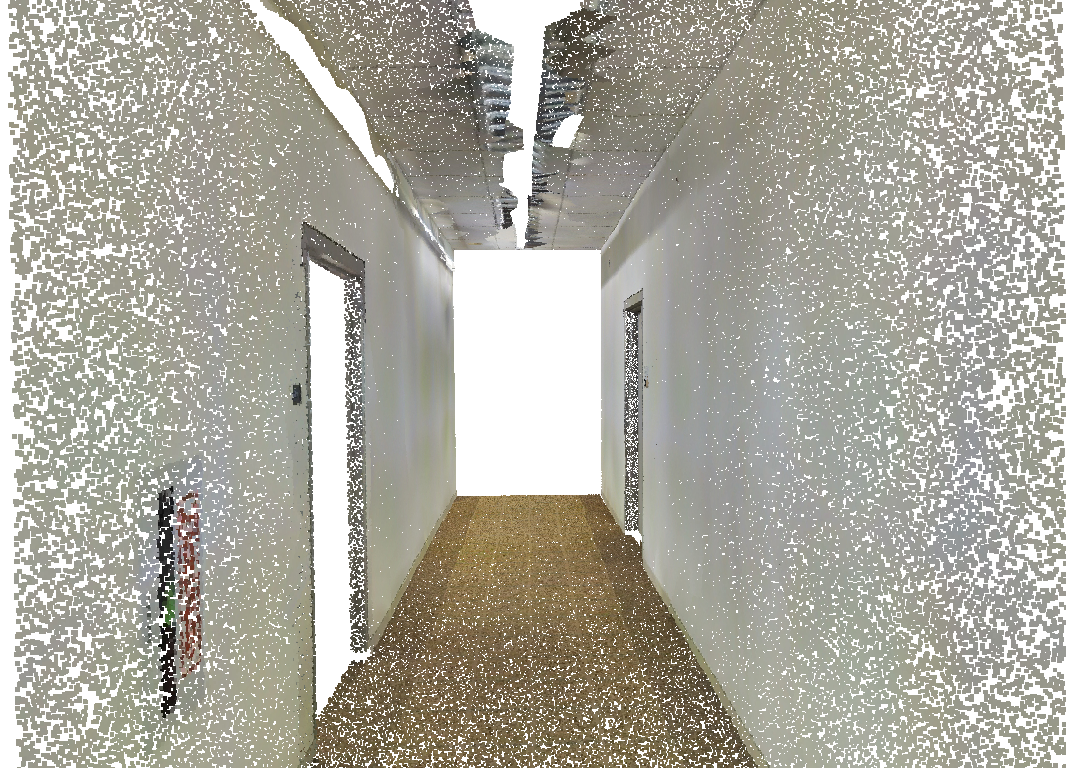}\\
		\rotatebox[origin=c]{90}{Ground Truth} &
		\includegraphics[align=c, height=0.13\linewidth]{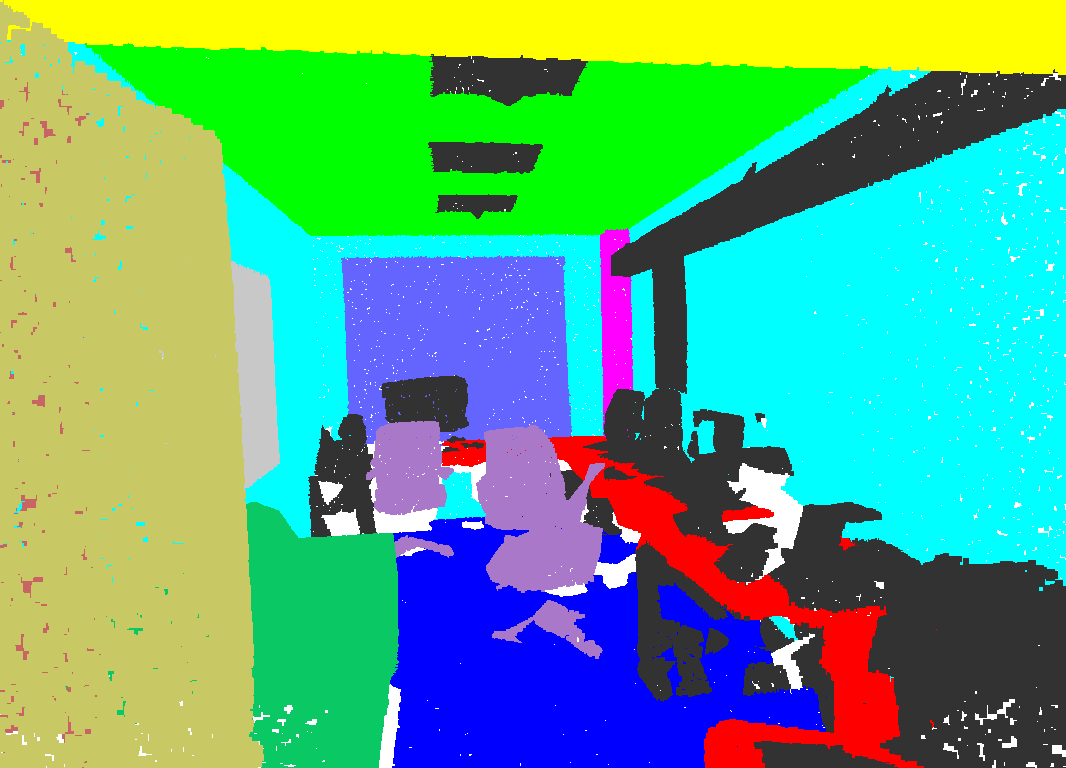}&
		\includegraphics[align=c, height=0.13\linewidth]{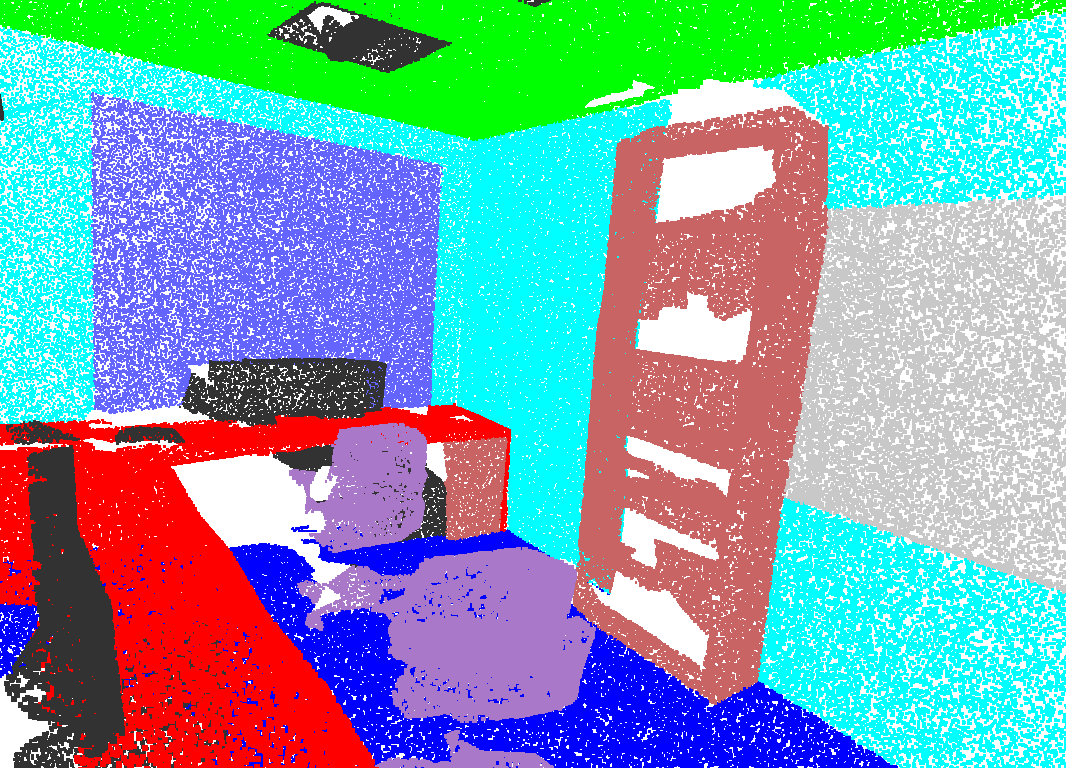}&
		\includegraphics[align=c, width=31mm, height=0.13\linewidth]{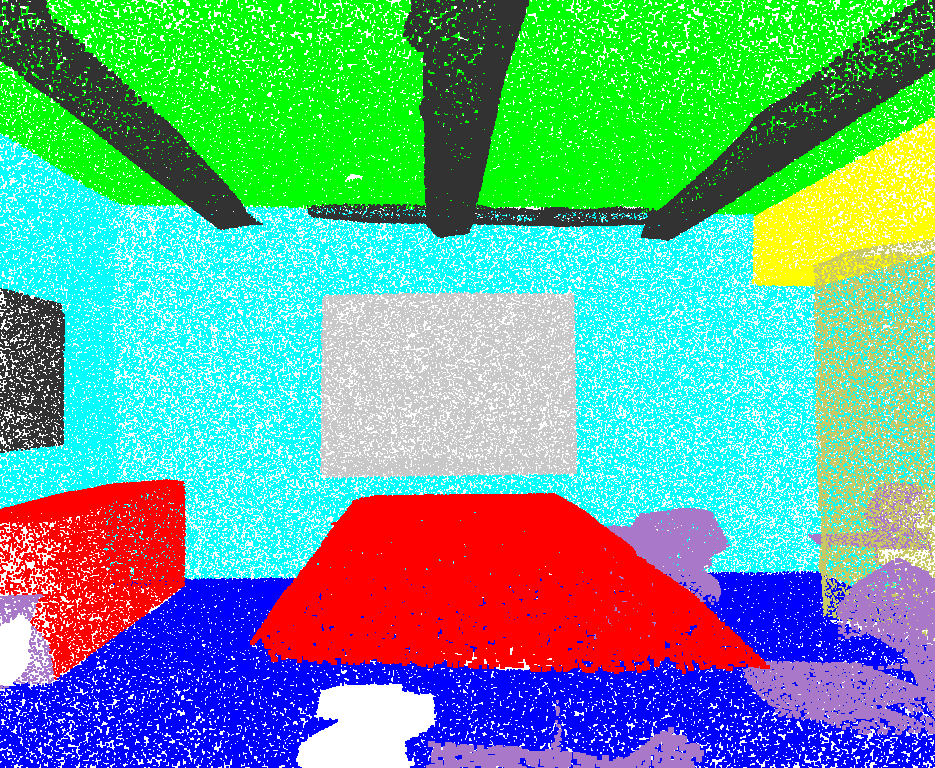}&
		\includegraphics[align=c, height=0.13\linewidth]{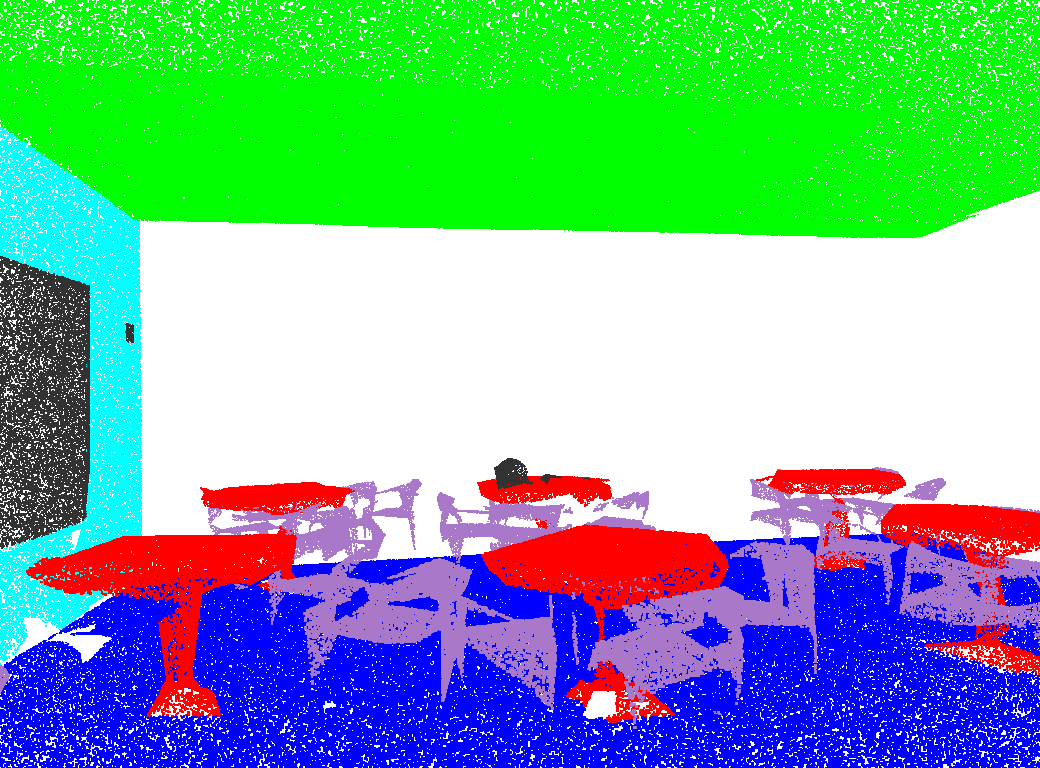}&
		\includegraphics[align=c, height=0.13\linewidth]{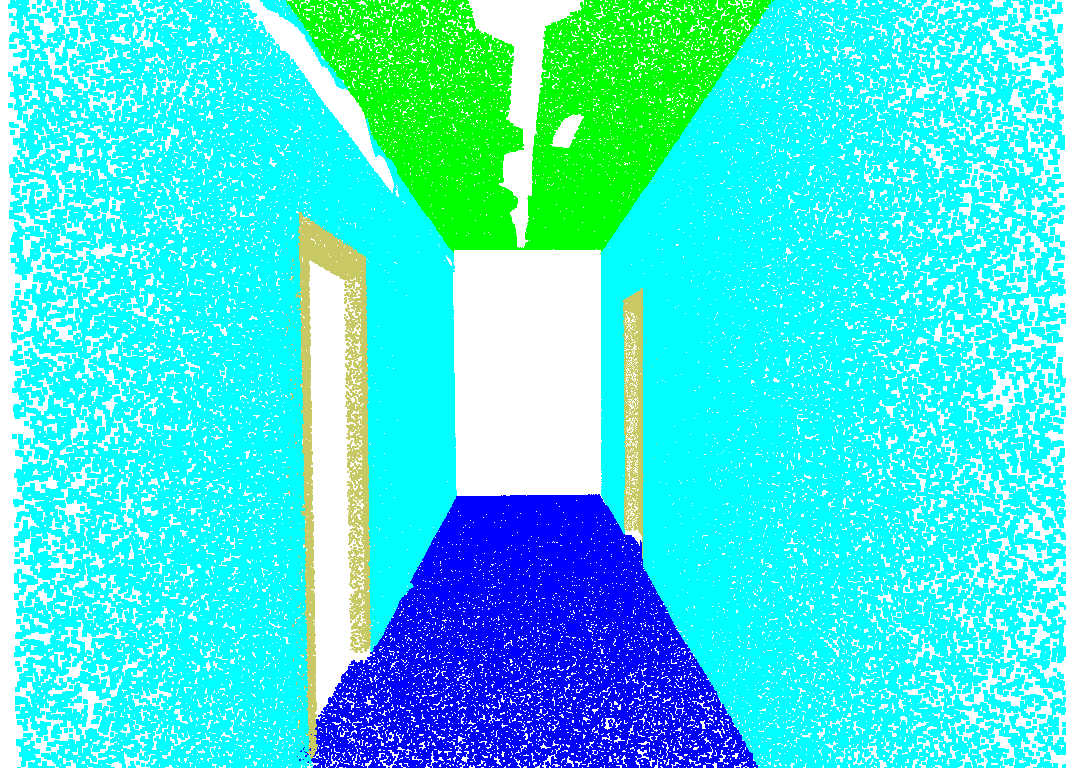}\\
		\rotatebox[origin=c]{90}{Ours} &
		\includegraphics[align=c, height=0.13\linewidth]{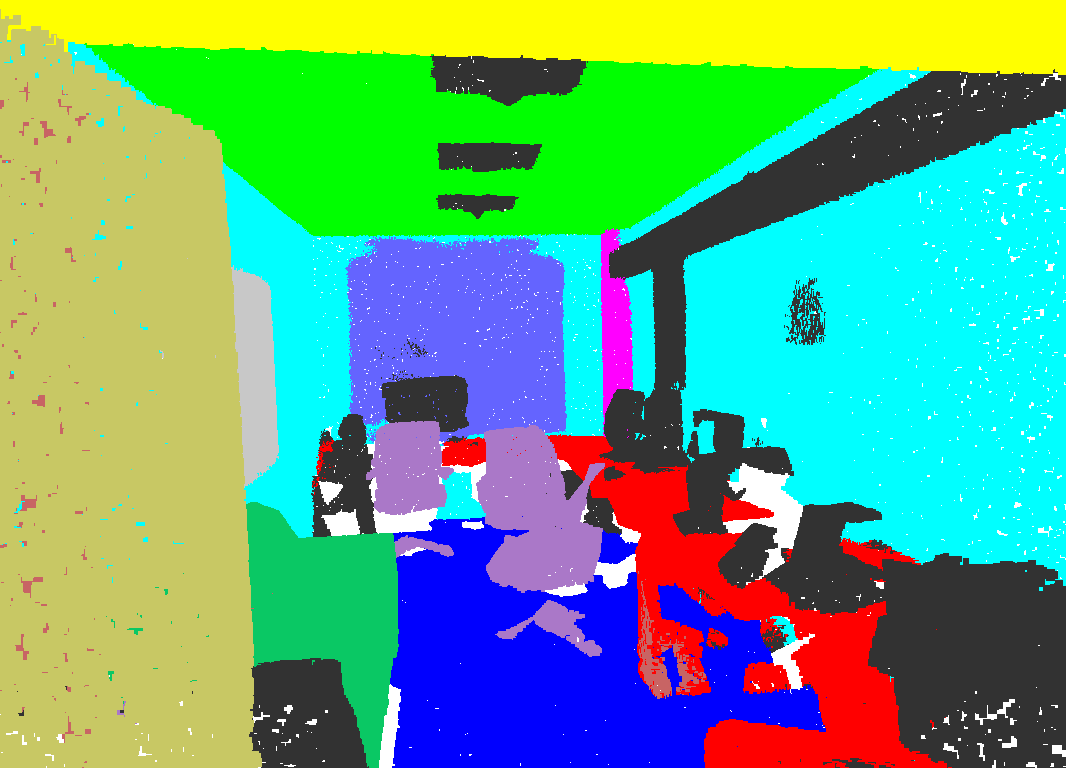}&
		\includegraphics[align=c, height=0.13\linewidth]{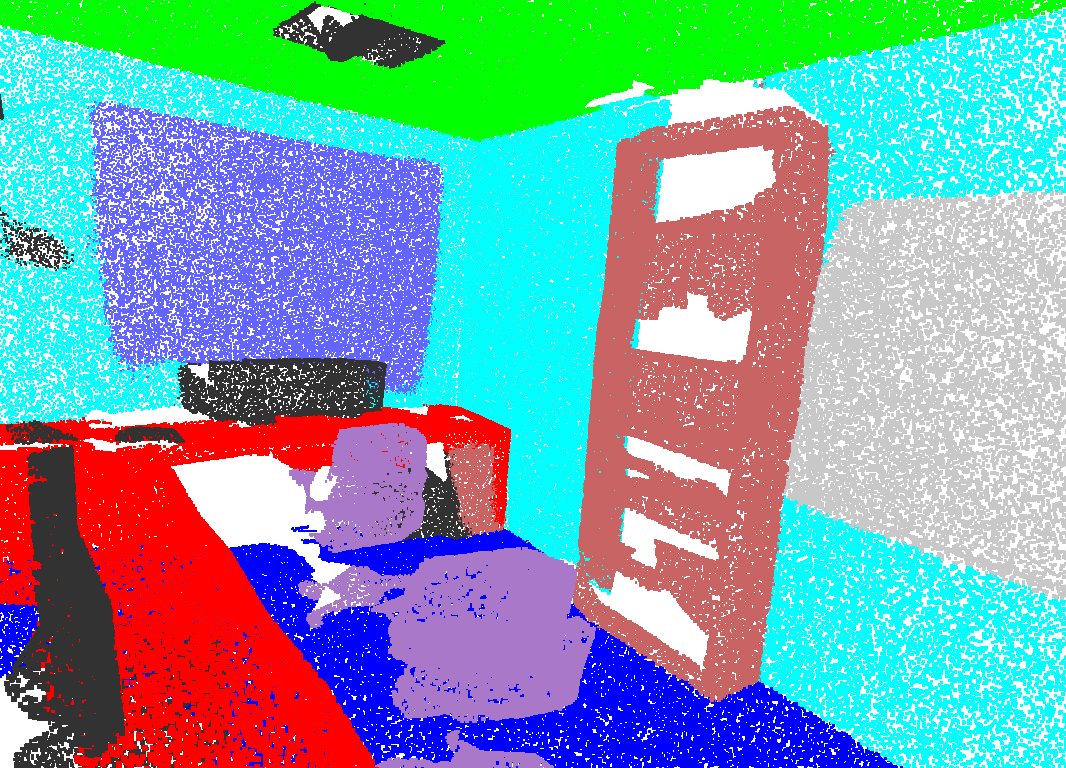}&
		\includegraphics[align=c, width=31mm, height=0.13\linewidth]{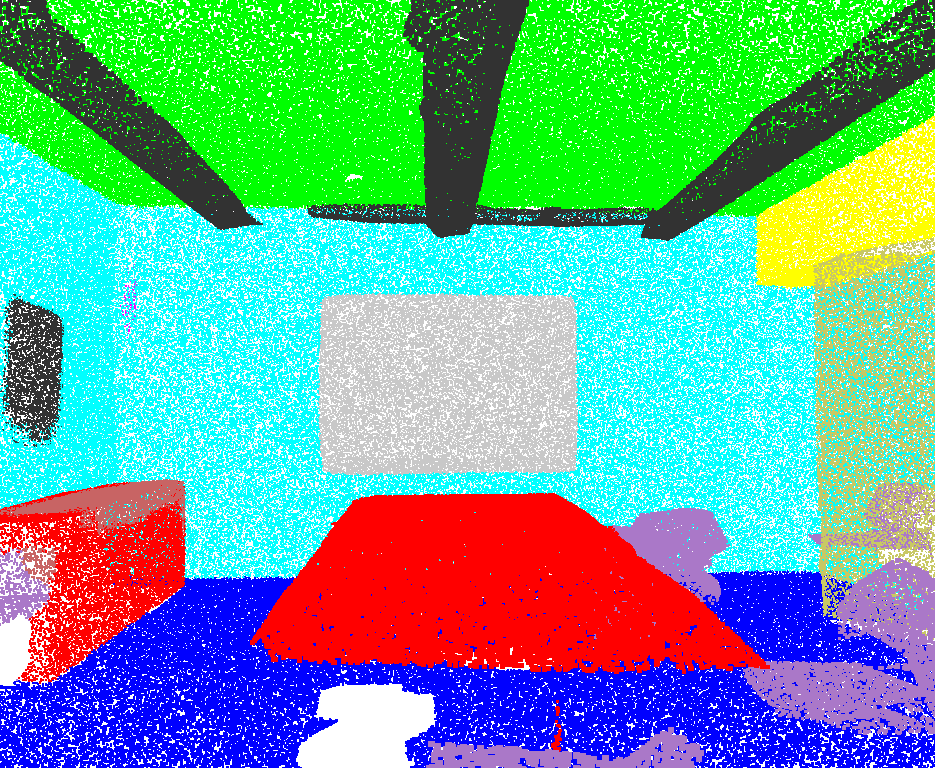}&
		\includegraphics[align=c, height=0.13\linewidth]{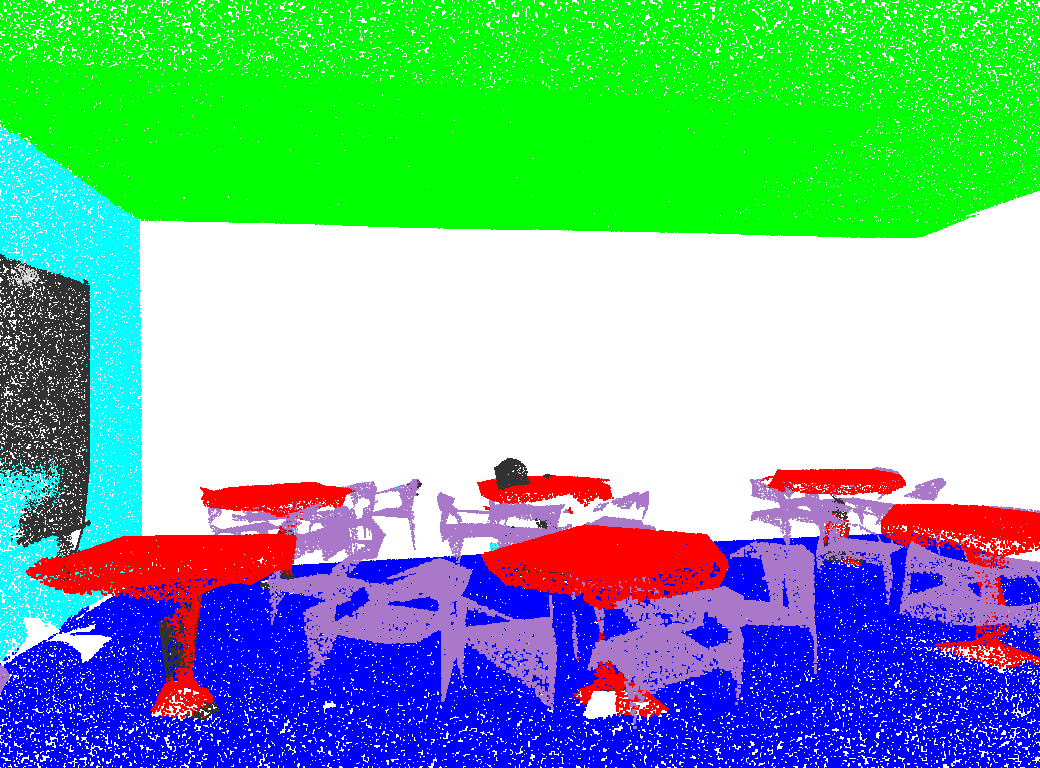}&
		\includegraphics[align=c, height=0.13\linewidth]{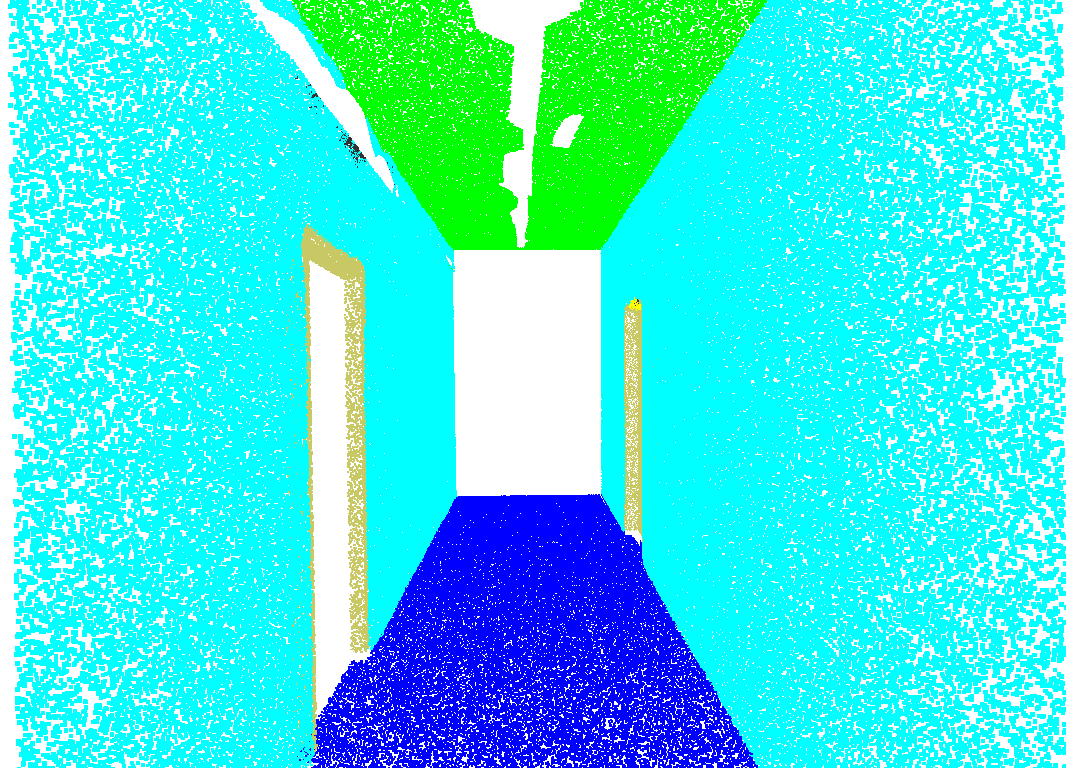}\\
	\end{tabular}
	}
	\vspace{5pt}
	\caption{Visualization of the semantic segmentation results on the S3DIS dataset.}
	\label{fig:s3dis}
	\vspace*{-1mm}
\end{figure*}

\subsection{Implementation Details}
The point branch contains an encoder with four down-sampling layers and a decoder with four upsampling layers.
The numbers of points, $N_0, N_1, N_2, N_3, N_4=N$, in the decoder are 
16, 64, 256, 1,024, and 4,096, respectively. The edge branch has five blocks with $k$ (number of nearest neighbors) set to $4, 6, 10, 14, 16$ from layer 0 to 4. $k$ is chosen as 3 for point and edge feature interpolation.

The whole network was trained in an end-to-end manner using the SGD optimizer with batch size 16 and base learning rate 0.05. For S3DIS, we train the network for 100 epochs and decay the rate by 0.1 for every 25 epochs. For ScanNet, we train the network for 120 epochs and decay the rate by 0.1 for every 30 epochs. The momentum and weight decay are set to 0.9 and 0.0001 respectively. 

\subsection{Datasets}

\paragraph{S3DIS}
The dataset~\cite{s3dis} has 6 areas with a total of 271 rooms.
Each room is provided as points with RGB information.
Each point has a semantic label from 13 categories of floor, window, door, etc.
In each training iteration, we randomly sample blocks in the training areas, with 4,096 points randomly selected per block. We set the block size as $0.8m \times 0.8m$ with $0.1m$ padding.
Also, we represent each point as a 9D vector with $XYZ$, $RGB$, and normalized position in room.
All points in the test areas are used in evaluation.
Two settings are adopted~\cite{segcloud, spg, pointcnn}: (i) splitting Area 5 as the test set and using others for training; and (ii) adopting 6-fold cross validation, with each of the 6 areas taking as the test set once. 

\vspace{-0.12in}
\paragraph{ScanNet v2} 
The dataset has 1,613 scans with a train/validation/test split of 1,201/312/100. Excluding the `unannotated' points, each point in the scans has a label from 20 categories of wall, shower curtain, etc. To prepare the input data, we follow previous work~\cite{pointnet2} to randomly sample blocks in rooms and sample 4,096 points per block.
Again, we use $0.8m \times 0.8m$ block size and $0.1m$ padding.
Here, each input point feature is a 6D vector ($XYZ$ \& RGB).
We evaluated on both the validation and test sets.
Since the semantic annotation for the test sets is not publicly available, we submitted our predictions to the official server to obtain the evaluation results.

\vspace{-0.12in}
\paragraph{Evaluation Metric}
It includes the class-wise mean of intersection over union (mIoU), class-wise mean of accuracy (mAcc) and point-wise overall accuracy (OA). 

\begin{figure*}
	\centering
	\begin{tabular}{@{\hspace{0.0mm}}c@{\hspace{1.0mm}}c@{\hspace{1.0mm}}c@{\hspace{1.0mm}}c@{\hspace{1.0mm}}c@{\hspace{1.0mm}}c@{\hspace{0.0mm}}}
		\rotatebox[origin=c]{90}{Input} &
		\includegraphics[align=c, width=0.18\linewidth]{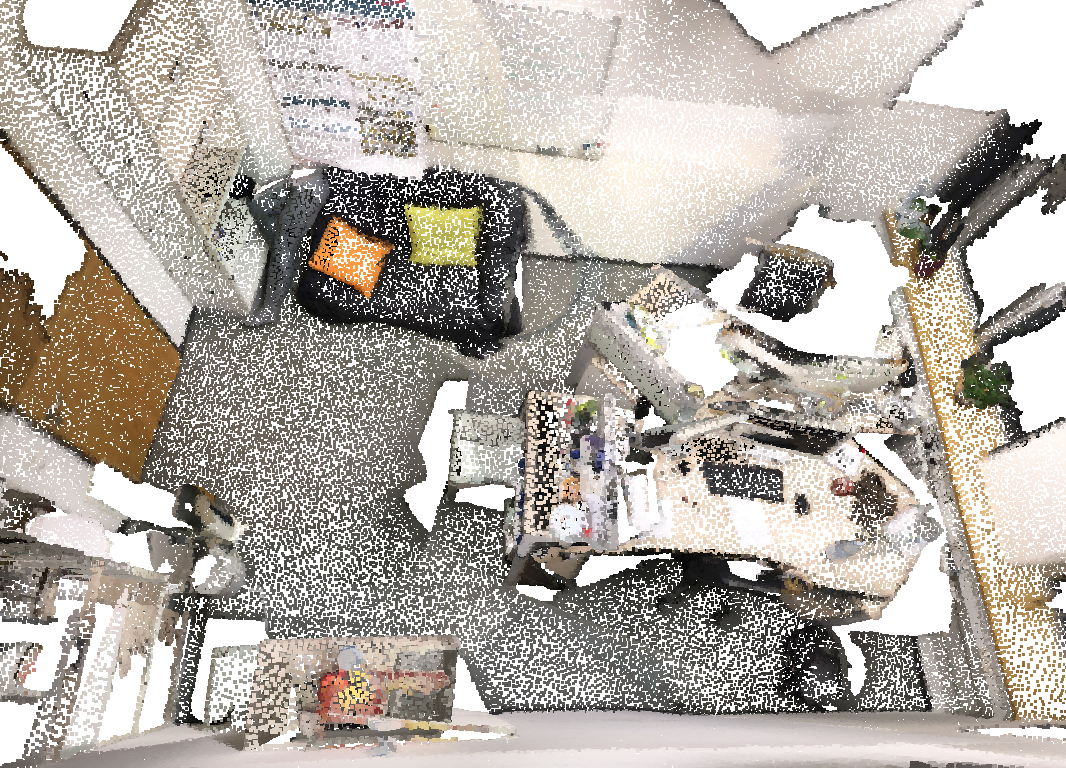}&
		\includegraphics[align=c, width=0.18\linewidth]{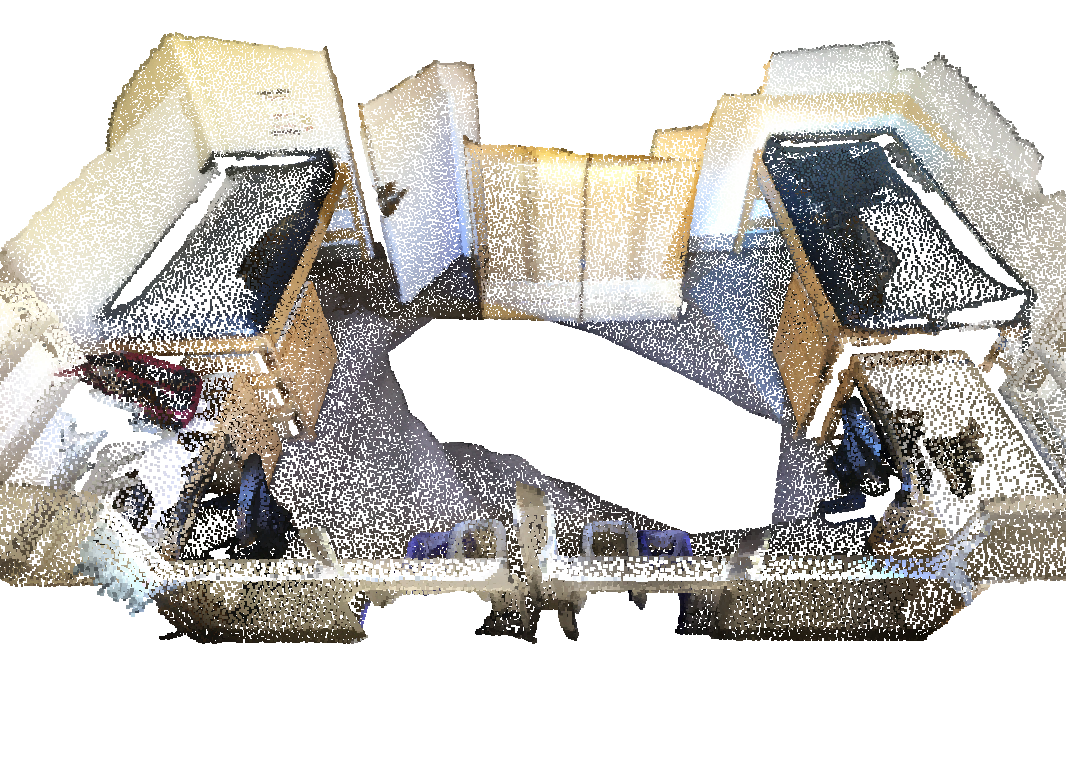}&
		\includegraphics[align=c, width=0.18\linewidth]{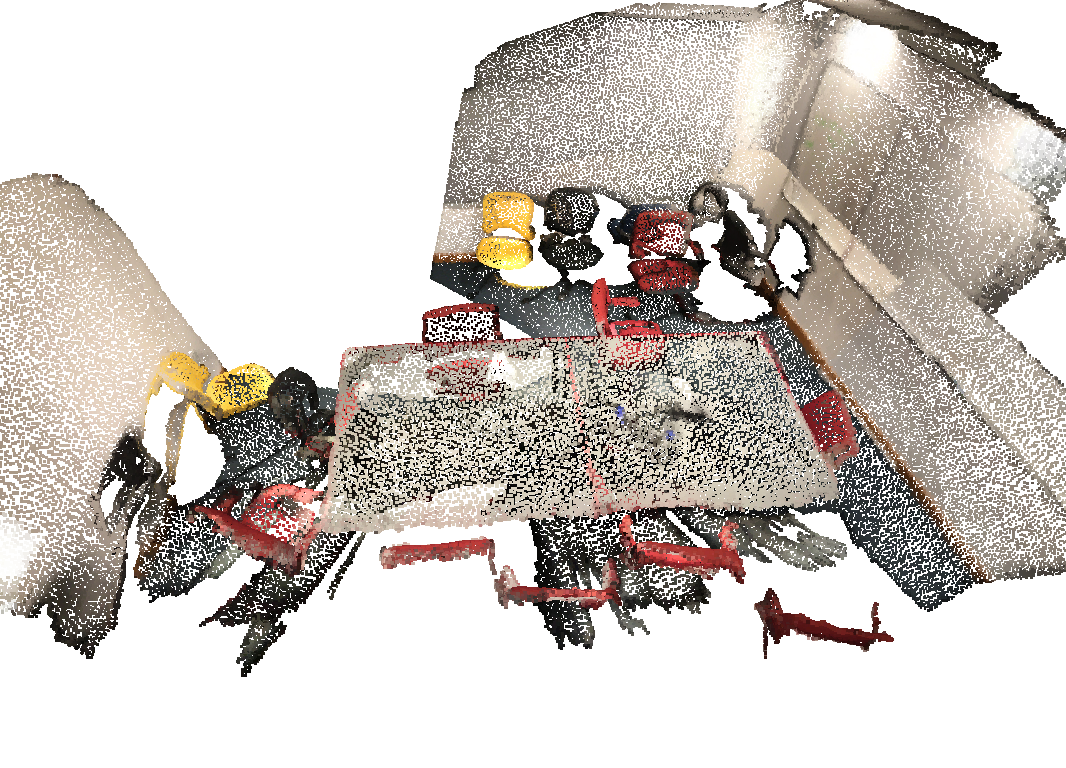}&
		\includegraphics[align=c, width=0.18\linewidth]{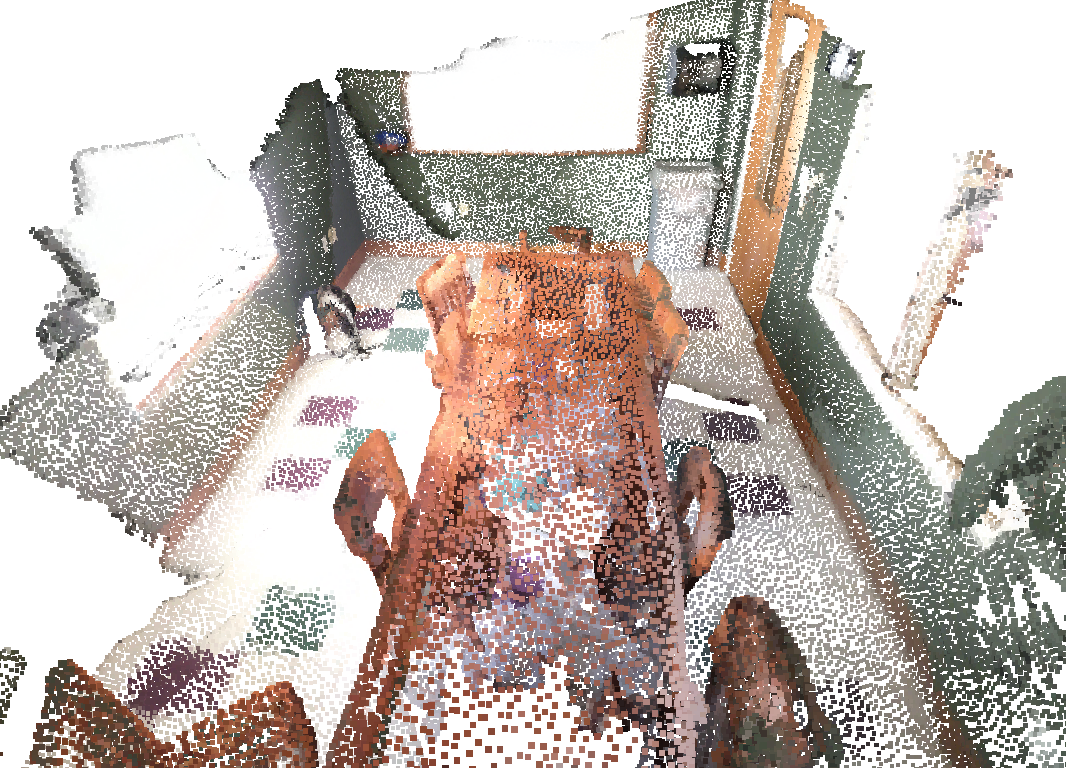}&
		\includegraphics[align=c, width=0.18\linewidth]{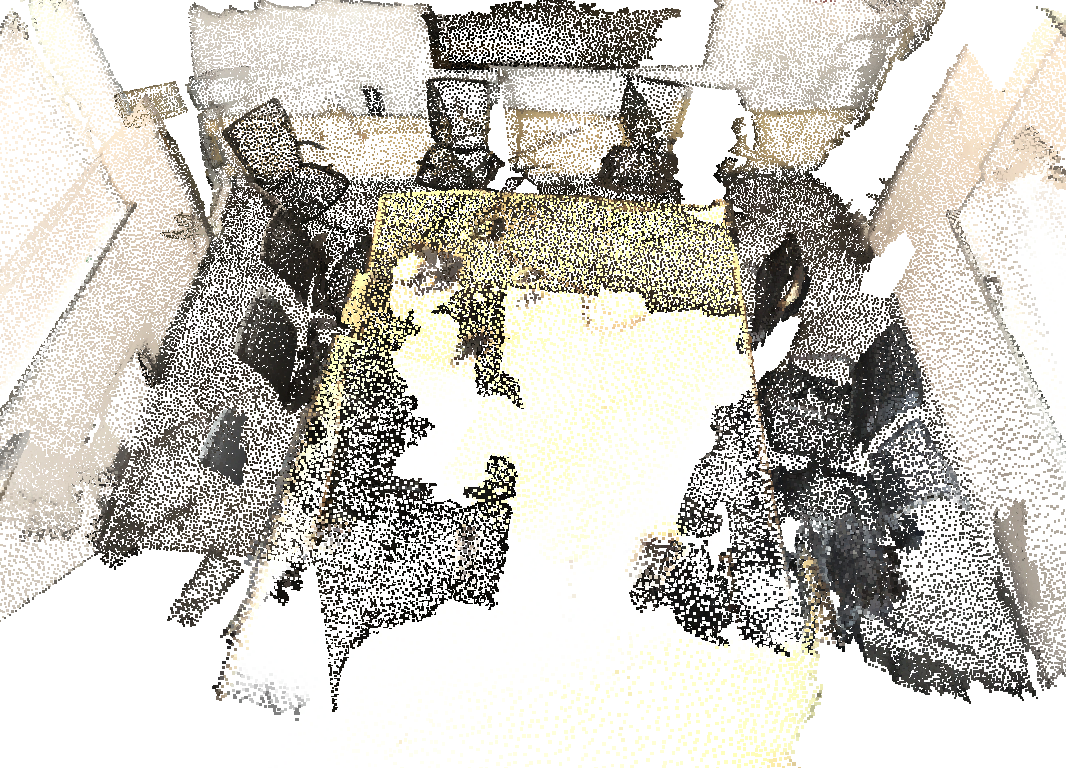}\\
		\rotatebox[origin=c]{90}{Ground Truth} &
		\includegraphics[align=c, width=0.18\linewidth]{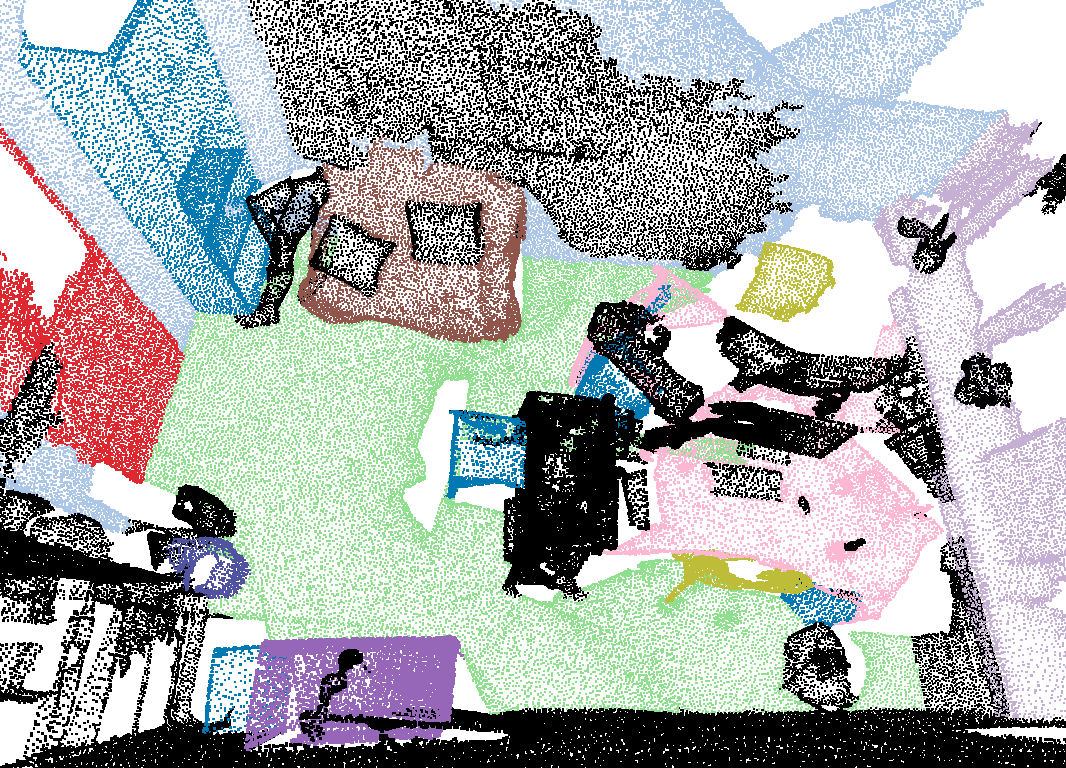}&
		\includegraphics[align=c, width=0.18\linewidth]{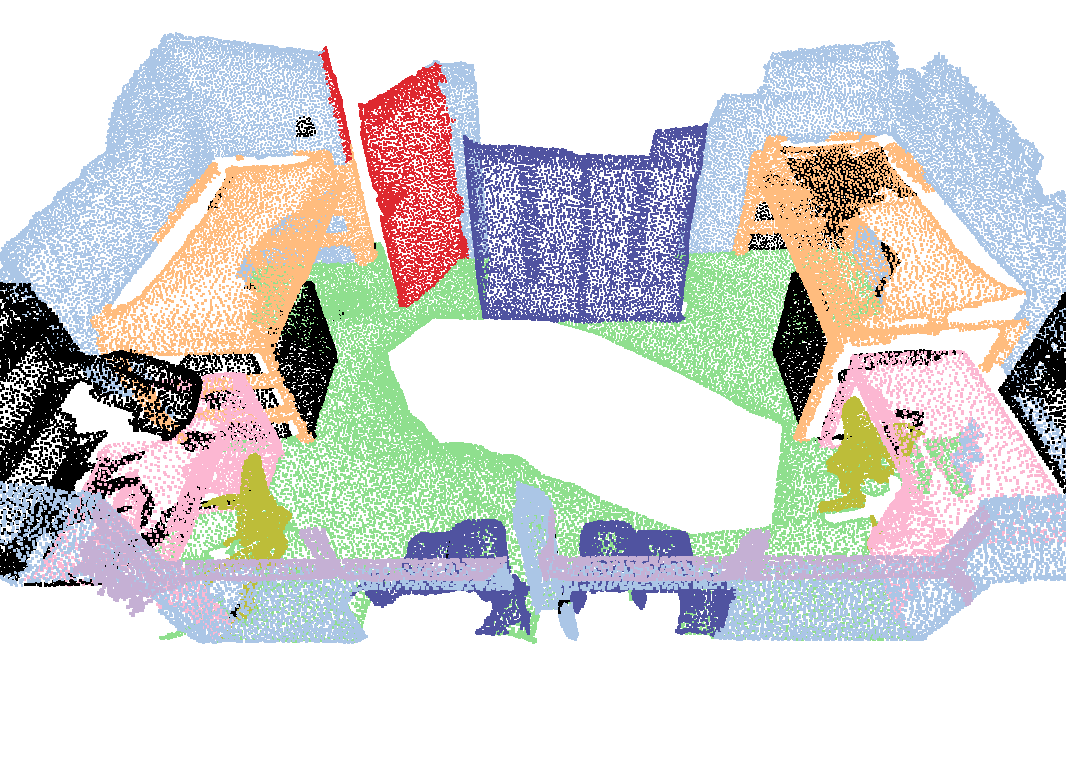}&
		\includegraphics[align=c, width=0.18\linewidth]{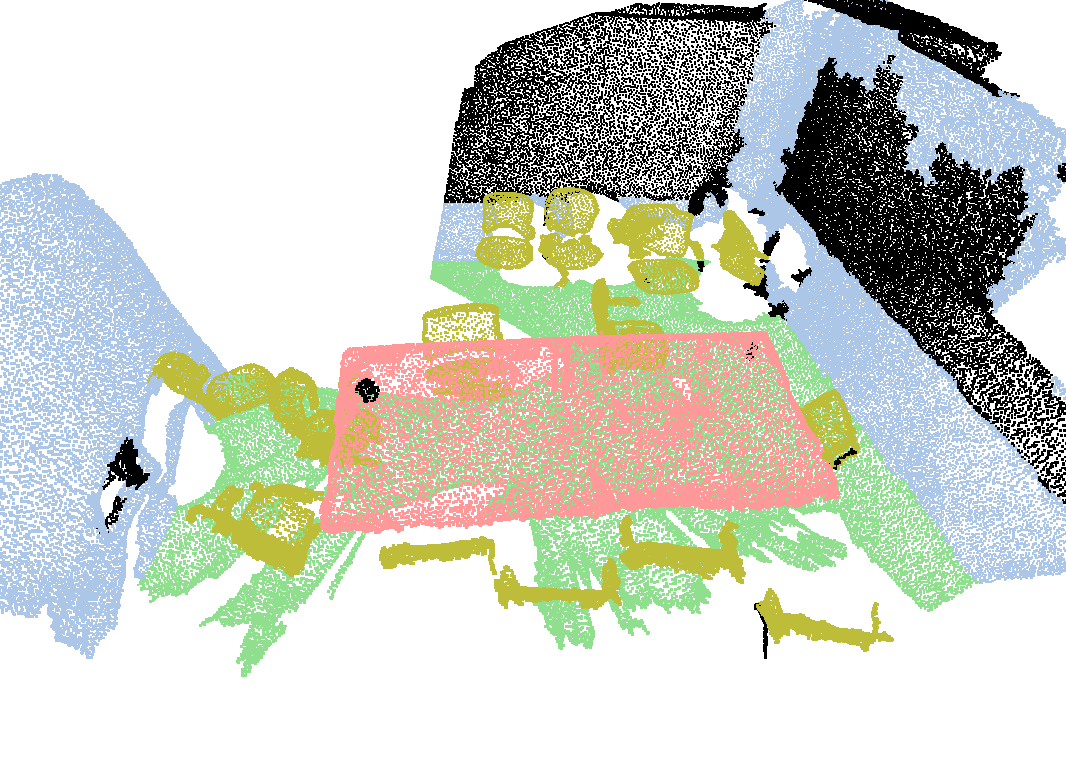}&
		\includegraphics[align=c, width=0.18\linewidth]{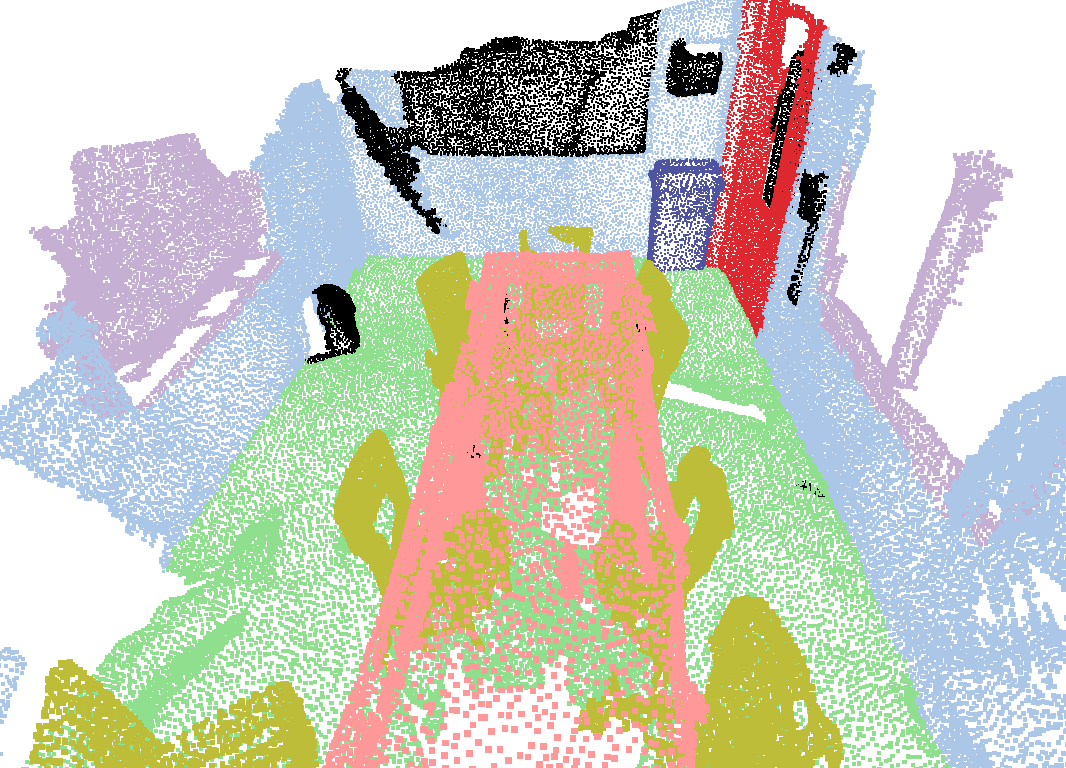}&
		\includegraphics[align=c, width=0.18\linewidth]{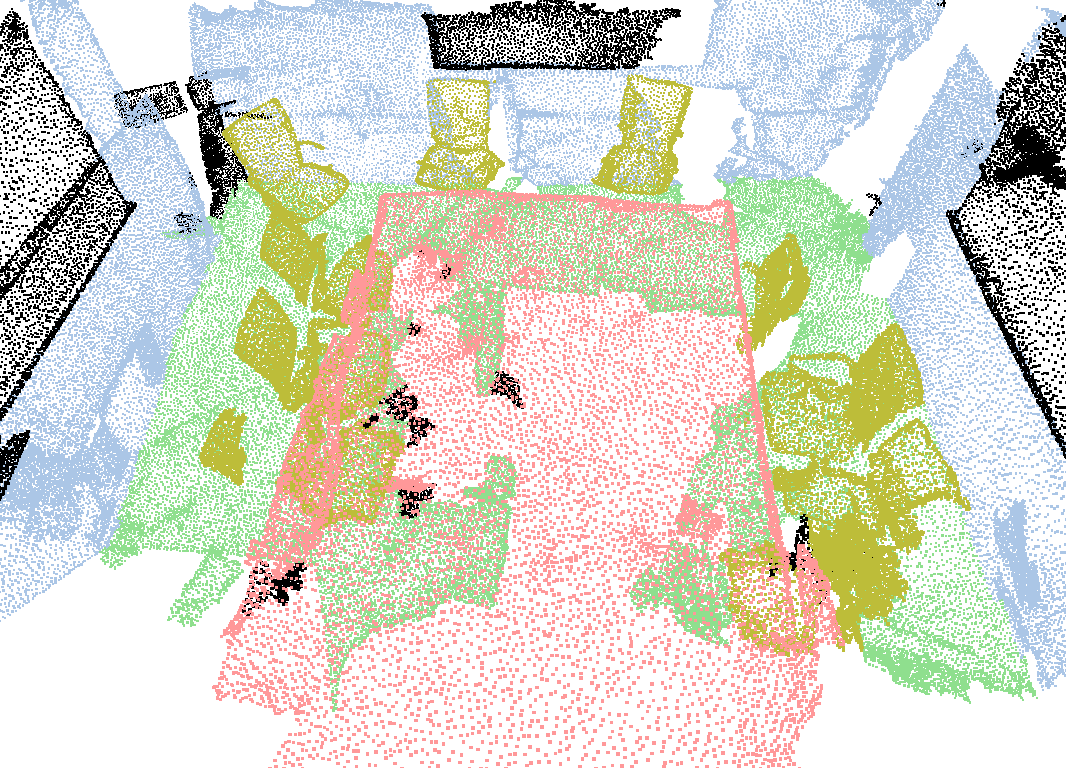}\\
		\rotatebox[origin=c]{90}{Ours} &
		\includegraphics[align=c, width=0.18\linewidth]{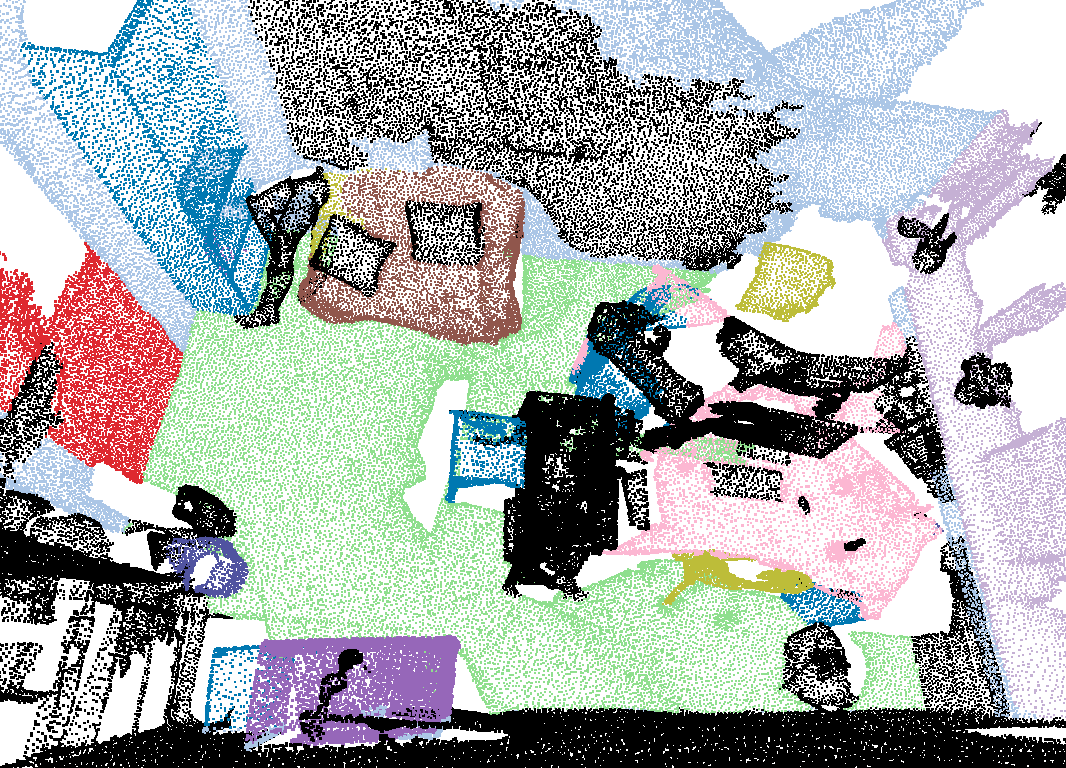}&
		\includegraphics[align=c, width=0.18\linewidth]{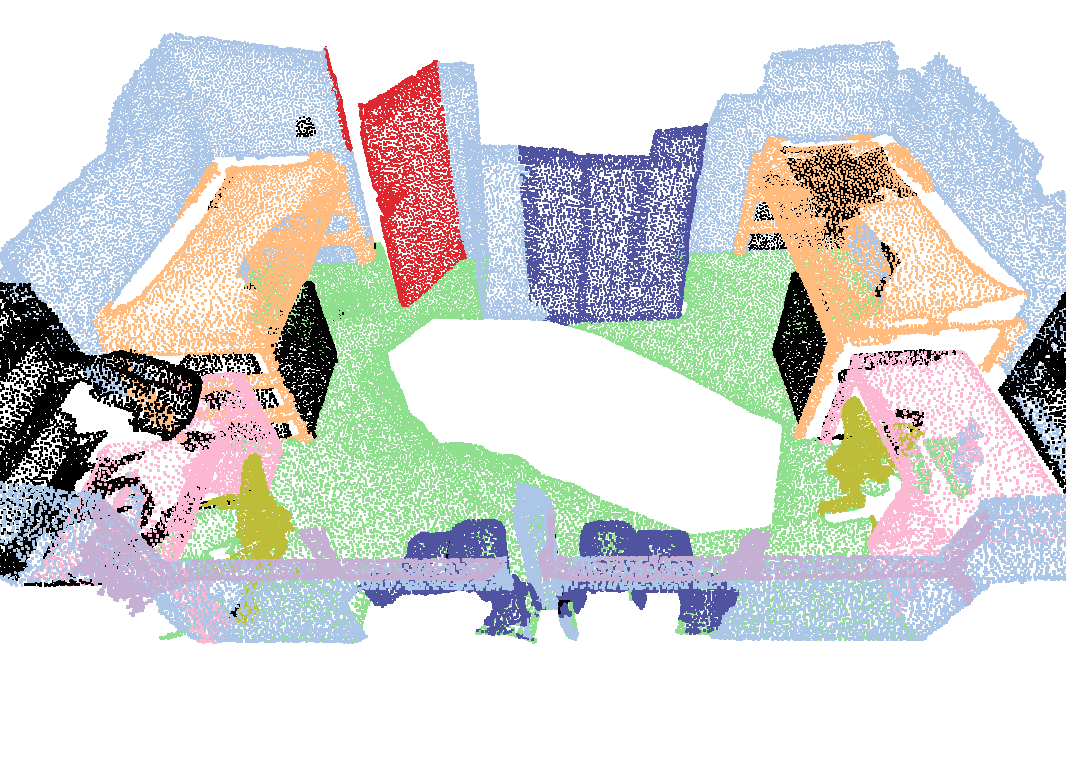}&
		\includegraphics[align=c, width=0.18\linewidth]{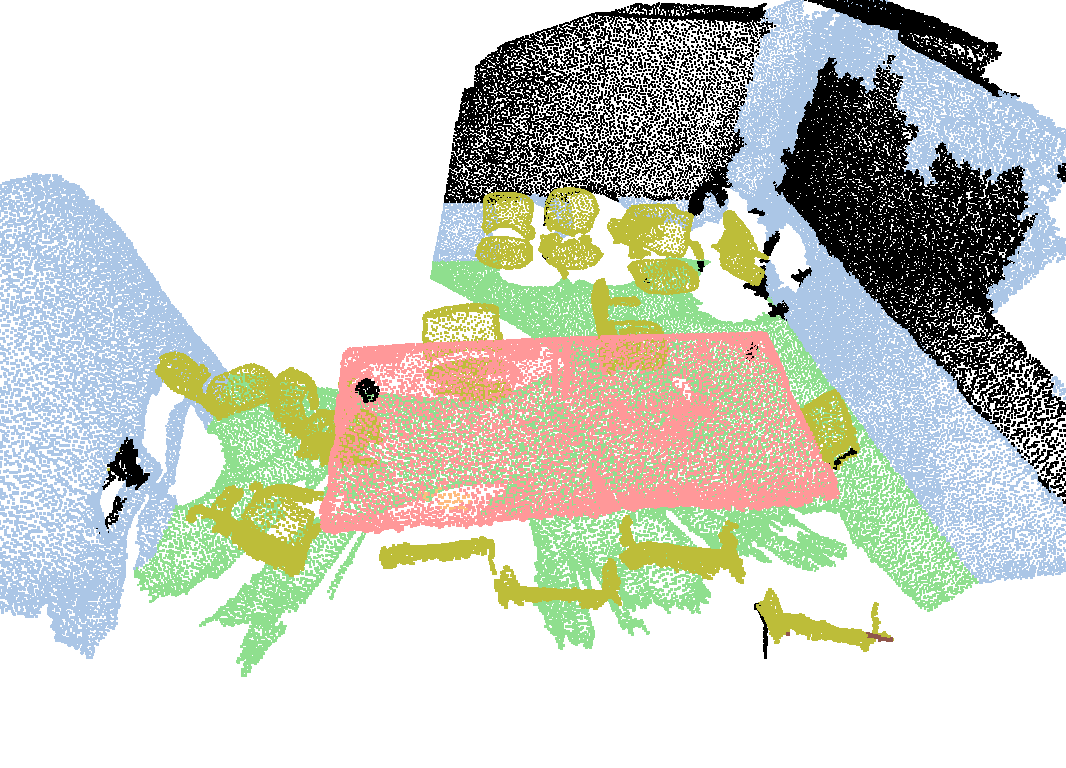}&
		\includegraphics[align=c, width=0.18\linewidth]{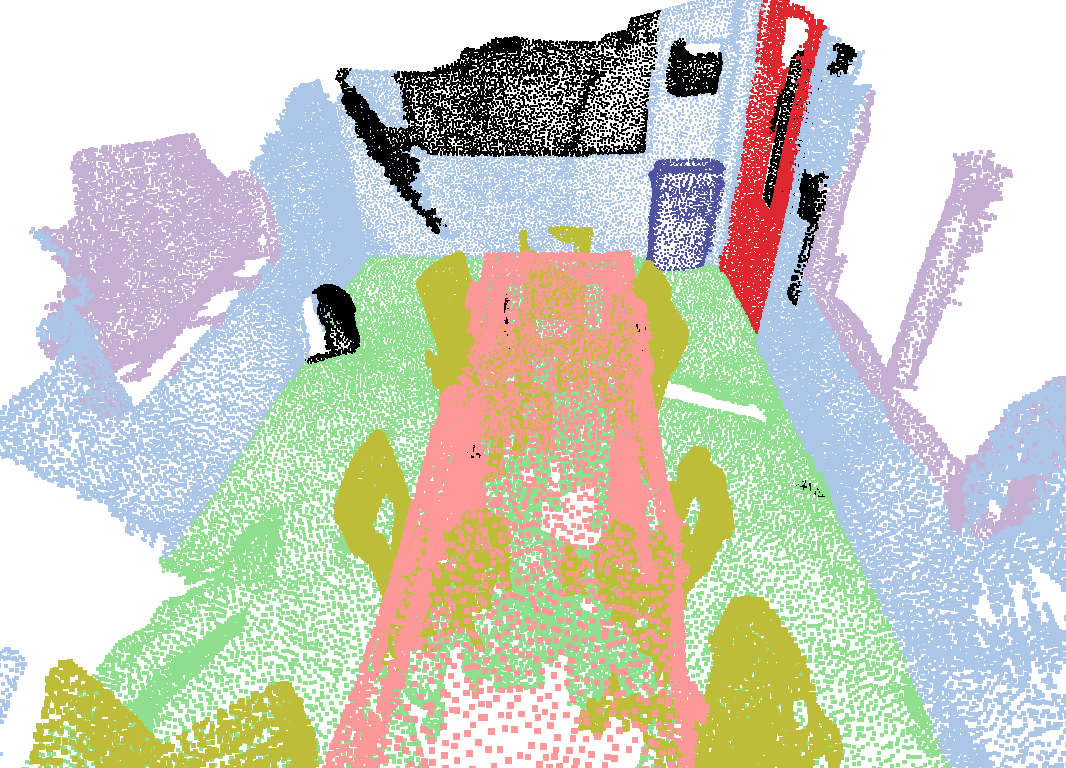}&
		\includegraphics[align=c, width=0.18\linewidth]{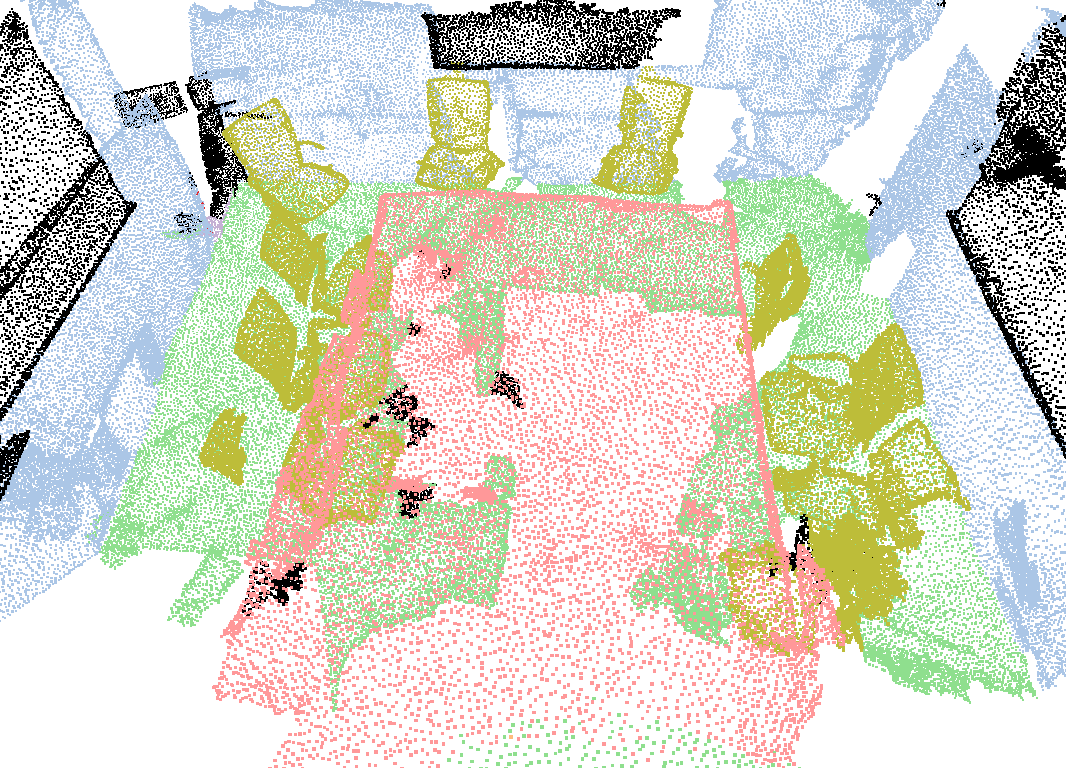}\\
	\end{tabular}
	\vspace{5pt}
	\caption{Visualization of the semantic segmentation results on the ScanNet v2 dataset.}
	\label{fig:scannet}
	\vspace*{-0mm}
\end{figure*}

\subsection{Main Results}
Table~\ref{tab:s3disresult} lists quantitative results of different methods on S3DIS Area 5.
Compared to previous approaches, ours yields the highest scores in terms of all the three metrics. Specifically, our model yields mIoU 61.85\%, exceeding the former best by 3.58\%.
Table~\ref{tab:s3disresult2} shows the comparison among different architectures on 6-fold cross validation. Ours also reaches the first place for all the three items.

Table~\ref{tab_scannetv2} lists results of our framework and other point-based methods on ScanNet v2 test set.
All methods use only point clouds with RGB color as input without voxelization.
Our approach outperforms others by a large margin: 6.2\% higher in absolute mIoU and 11.2\% better relatively. Visual results are shown in Figs.~\ref{fig:s3dis} and~\ref{fig:scannet}.
Our method segments objects even in complex scenes.
It is notable that several detailed structures are classified and segmented from the surroundings, manifesting the effectiveness of our method.

\begin{table}[t]
	\small
	\setlength{\tabcolsep}{5pt}
	\begin{center}
		\begin{tabular}{ l | c c c}
			\toprule[1.0pt]
			Methods & OA & mAcc & mIoU\\
			\hline
			PointNet~\cite{pointnet} & 78.5 & 66.2 & 47.6\\
			RSNet~\cite{huang2018recurrent} & - & 66.45 & 56.47\\
			SPGraph~\cite{spg} & 85.5 & 73.0 & 62.1\\
			PointCNN~\cite{pointcnn} & 88.14 & 75.61 & 65.39\\
			\hline
			Our Method & \textbf{88.20} & \textbf{76.26} & \textbf{67.83}\\
			\bottomrule[1.0pt]
		\end{tabular}
	\end{center}
	\vspace*{-1mm}
	\caption{Semantic segmentation results on the S3DIS dataset with 6-fold cross validation.}
	\label{tab:s3disresult2}
	\vspace*{-2mm}
\end{table}

\begin{table}
	\setlength{\tabcolsep}{8pt}
	\begin{center}
		\begin{tabular}{ l | c }
			\toprule[1.0pt]
			Method & mIoU  \\
			\hline
			PointNet++~\cite{pointnet2} & 33.9 \\
			SPLATNet~\cite{splatnet} & 39.3  \\
			PointCNN~\cite{pointcnn} & 45.8 \\
			PointConv~\cite{pointconv} & 55.6  \\
			\hline
			Our Method & \textbf{61.8} \\
			\bottomrule[1.0pt]
		\end{tabular}
	\end{center}
	\vspace{-1mm}
	\caption{Semantic segmentation results on ScanNet v2 test set.}
	\vspace{-2mm}
	\label{tab_scannetv2}
\end{table}

\subsection{Ablation Study}
For ScanNet v2, the models are trained on training set and evaluated on validation set. For S3DIS, the models are trained on Areas 1-4 \& 6 and evaluated on Area 5.   

\vspace{-3mm}
\paragraph{Edge Function}
We explore different ways of incorporating point information into edges, including Subtraction, Summation, Hadamard product, `ConcatSub', and Concatenation. Here `ConcatSub' is defined as 
\begin{equation}
f_{edge}(F_i^{L}, F_j^{L}) =   [(p_j - p_i), \text{ } F_j^{L}, \text{ }(F_j^{L} -  F_i^{L})].
\end{equation}
Table~\ref{tab:ablation_edgefunction} shows comparison of the results.
Overall, concatenation yields the best result due to preservation of most point information.
Summation, Subtraction, and Hadamard Product all cause information loss in the level of point features. `ConcatSub' achieves similar performance with Concatenation, since the two-point features can be restored in this type of operations.

\begin{table}
	\setlength{\tabcolsep}{7pt}
	\begin{center}
		\resizebox{0.9\linewidth}{!}{%
		\begin{tabular}{ l | c c c }
			\toprule[1.0pt]
			Methods & mIoU  & mAcc & OA   \\
			\hline
			Subtraction &  58.31 / 58.85 & 67.95 / 65.66 & 84.02 / 86.44 \\
			Summation & 57.86 / 58.96 & 67.25 / 65.87 & 83.69 / 86.48  \\
			Hadamard Product & 59.07 / 58.79 & 68.02 / 65.27 & 85.31 / 86.16 \\
			ConcatSub & 63.09 / 59.37 & 71.82 / 66.19 & 86.12 / 86.53\\
			\hline
			Concatenation & \textbf{63.36} / \textbf{61.85} & \textbf{72.61} /  \textbf{68.30} & \textbf{86.13} /  \textbf{87.18}  \\
			\bottomrule[1.0pt]
		\end{tabular}
		}
	\end{center}
	\caption{Ablation study results for edge function $f_{edge}$ on ScanNet v2 and S3DIS. The results are shown in format of ScanNet v2 / S3DIS. The ablation on two datasets share similar observation.\vspace{2mm}}
	\label{tab:ablation_edgefunction}
\end{table}

\begin{table}
	\setlength{\tabcolsep}{5pt}
	\begin{center}
	\resizebox{0.9\linewidth}{!}{%
		\begin{tabular}{ l | c c c }
			\toprule[1.0pt]
			Methods & mIoU  & mAcc & OA   \\
			\hline
			AdaAggre (w. softmax) &  56.44 & 66.17 & 83.06 \\
			AdaAggre (w.o. softmax) & 55.01 & 64.12 & 82.67  \\
			\hline
			MaxPool + Concat & \textbf{63.36} & \textbf{72.61} & \textbf{86.13} \\
			\bottomrule[1.0pt]
		\end{tabular}
	}
	\end{center}
	\caption{Ablation results for message passing by edges.}
	\vspace{-2mm}
	\label{tab_ablation_edgepass}
\end{table}

\vspace{-3mm}
\paragraph{Message Passing by Edges}
\label{sec_edge_usage}
Besides the approach described in Section~\ref{sec_interaction}, we also experimented with another scheme which is inspired by graph convolution~\cite{kipf2016semi, velivckovic2017graph}, where the edge features are further encoded to form weights for the linked points. The point features are then updated as a weighted sum of the adjacent point features. We denote this scheme as adaptive aggregation (\textit{AdaAggre}) and test the two settings, with and without softmax, for the weights.
Table~\ref{tab_ablation_edgepass} lists the experimental results on ScanNet v2 validation set.

The performance gain for the graph-convolution-style methods is lower than max-pooling followed by concatenation. It may be because during the point decoding, it is not very helpful to mix point features in each local neighborhood. 
Instead, the combined contextual feature reveals the relation of a point with its neighborhood. It can better preserve the point's own distinctiveness.

\vspace{-3mm}
\paragraph{Hierarchical Graph Construction and Edge Upsampling}
We build connection between edge features of adjacent layers by ``edge upsample''.
We also experimented on ScanNet dataset with removing hierarchical graph construction and
building the graph of each layer separately without edge upsampling.

The mIoU/mAcc/OA (\%) results are 57.01/66.52/83.57 respectively, much lower than our full framework with 63.36/72.61/86.13. The connected edge branch optimally incorporates the point features in different layers, enabling effective learning for the edge features. 

\vspace{-1mm}
\section{Conclusion}
We have designed a hierarchical point-edge interaction network, in which an edge branch is proposed to work with the encoder-decoder point branch for point cloud semantic segmentation. The proposed hierarchical graph framework enables the edge branch to progressively integrate different-layer point features. Also, the generated edge features are incorporated into the point branch to provide contextual information. The final edge features are supervised by the semantic consistency of related points to implicitly regularize the point features. All these steps make semantic relationship with local context well utilized via edges.

With the high-quality point prediction results and generality of the framework applicable to different datasets, we believe the proposed method will broadly benefit 3D understanding in the community. In the future, we will explore multi-range edge construction to gather both close-range and long-distance contextual information.

\paragraph{Acknowledgments}
This project is supported in part by the Research Grants Council of the Hong Kong Special Administrative Region (CUHK 14203416 \& 14201918).

{\small
	\bibliographystyle{ieee_fullname}
	\bibliography{egbib}
}

\end{document}